\pdfoutput=1

\documentclass[11pt]{article}

\usepackage[]{acl}

\usepackage{times}
\usepackage{latexsym}

\usepackage[T1]{fontenc}

\usepackage[utf8]{inputenc}

\usepackage{microtype}

\usepackage{inconsolata}
\usepackage{amsmath}
\usepackage{bbm}
\usepackage{amsfonts} 

\usepackage{enumitem}

\DeclareMathOperator*{\argmax}{arg\,max}

\usepackage{graphicx}
\usepackage{listings}
\usepackage{subcaption} 
\usepackage{booktabs}
\usepackage{xcolor}
\usepackage{multirow}
\usepackage{multicol}
\usepackage{centernot}
\usepackage{tabularx}
\usepackage{pgfplots}
\pgfplotsset{compat=1.18}
\usepackage{pifont}
\usepackage{tikz}
\usepackage{cleveref}
\usepackage{amssymb}
\usepackage{fdsymbol}
\newcommand{\minisection}[1]{\noindent{\bf #1}\hspace{0.6em}}

\title{Calibrating Large Language Models with Sample Consistency}

\author{%
  Qing Lyu $^{ * \spadesuit}$ \quad Kumar Shridhar $^{ * \blacklozenge}$\  \quad Chaitanya Malaviya $^{\spadesuit}$ \quad Li Zhang $^{\spadesuit}$  \quad Yanai Elazar $^{\clubsuit}$$^{\varheartsuit}$\\ \quad \bf{Niket Tandon} $^{\clubsuit}$ \quad \bf{Marianna Apidianaki} $^{\spadesuit}$ \quad  \bf{Mrinmaya Sachan} $^{\blacklozenge}$ \quad \bf{Chris Callison-Burch} $^{\spadesuit}$$^{\clubsuit}$ \\
  \\
 $^{\spadesuit}$ University of Pennsylvania \quad $^{\blacklozenge}$ ETH Zurich \quad \\ $^{\varheartsuit}$ University of Washington \quad $^{\clubsuit}$  Allen Institute for AI \\
 \\
 \texttt{lyuqing@sas.upenn.edu} \quad \texttt{shkumar@ethz.ch} 
  }

\newcommand\blfootnote[1]{%
  \begingroup
  \renewcommand\thefootnote{}\footnote{#1}%
  \addtocounter{footnote}{-1}%
  \endgroup
}
  
\begin{document}
\maketitle
\begin{abstract}


Accurately gauging the confidence level of Large Language Models' (LLMs) predictions is pivotal for their reliable application. However, LLMs are often uncalibrated inherently and elude conventional calibration techniques due to their proprietary nature and massive scale.
In this work, we explore the potential of deriving confidence from the distribution of multiple randomly sampled model generations, via three measures of \emph{consistency}. We perform an extensive evaluation across various open and closed-source models on nine reasoning datasets. Results show that consistency-based calibration methods outperform existing post-hoc approaches. Meanwhile, we find that factors such as intermediate explanations, model scaling, and larger sample sizes enhance calibration, while instruction-tuning makes calibration more difficult. Moreover, confidence scores obtained from consistency have the potential to enhance model performance. Finally, we offer practical guidance on choosing suitable consistency metrics for calibration, tailored to the characteristics of various LMs.
 \blfootnote{$*$ Equal contribution; Qing Lyu did her work while interning at Allen Institute for AI.}
 \footnote{Code:\url{https://github.com/veronica320/Calibrating-LLMs-with-Consistency}}
\end{abstract}

\section{Introduction}


Large Language Models (LLMs) excel in various tasks, yet it is hard to know when they err. 
A first step towards making LLMs more trustworthy is for them to provide a confidence estimate with predictions \cite{papadopoulos2001confidence}.
This estimate needs to be \emph{calibrated}, meaning that the confidence level is aligned with the likelihood of the prediction being correct
\cite{brier1950verification}. A well-calibrated system can enable model developers to provide selective predictions, help users decide when to trust or distrust model responses, and potentially facilitate performance improvement through human intervention or self-refinement \cite{self-refine, shridhar2023screws}.

\begin{figure}[t!]
\centering
\includegraphics[width=\columnwidth]{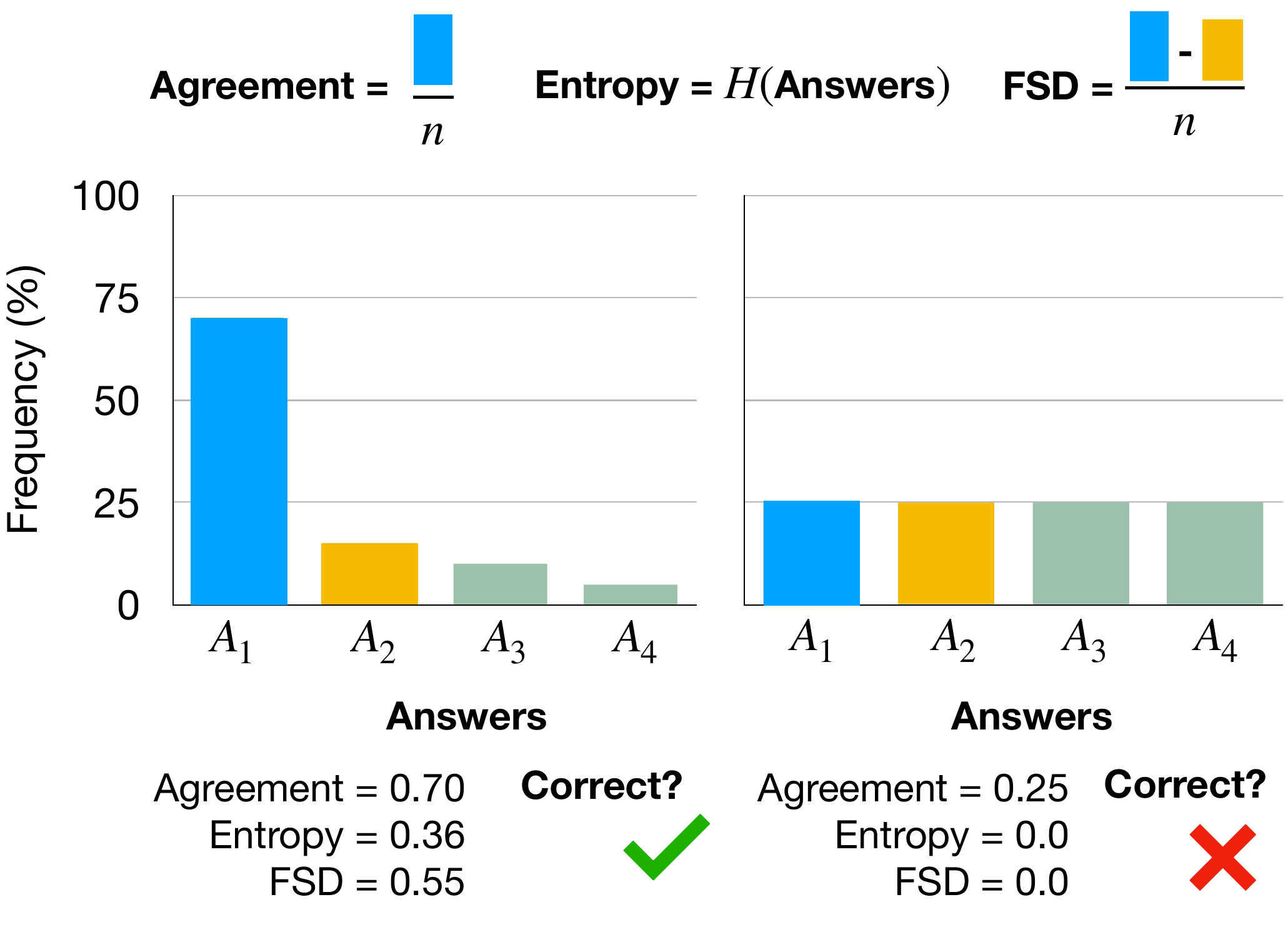}
\caption{We study three consistency measures in this work: agreement-based, entropy-based, and first-second-distance-based (FSD). Higher consistency suggests a higher likelihood of correctness, and vice versa.}
\vspace{-0.13in}
\label{fig:main_figure}
\end{figure}

Unfortunately, LLMs are not well-calibrated off-the-shelf --- the probability logits of model predictions are often poorly aligned with actual performance \cite{jiang_how_2021, chen_close_2023}. While traditional calibration methods \cite[][i.a.]{calibration-guo, lakshminarayanan_simple_2017, gal_dropout_2016} can in theory be used to better calibrate open-source LMs, 
for recent LLMs, these methods sometimes become formidably costly because of the need to retrain multiple copies of the model, and might even be inapplicable due to inaccessible training data, model weights, and output probabilities in closed-source LLMs.

In light of these issues, a promising recent line of work measures the \textit{consistency} of model generations to calibrate confidence \cite[][i.a.]{selfconsistency, xiong2023can}, with the advantage of being fully post-hoc and requiring no additional calibration data. 
However,
existing work has only used the \textit{agreement} between the original generation and multiple randomly sampled generations as a metric for consistency, ignoring the rich information from the \textit{distribution} of generations. \autoref{fig:main_figure} (right) shows an example of how agreement can potentially lead to overconfidence only based on one most popular answer, when all the answers are equally frequent. This creates a need for better consistency measures, and a systematic empirical comparison of them.

In this work, we investigate the research question: \emph{How can we best elicit a model's confidence from the consistency of multiple generations?} As shown in Figure~\ref{fig:main_figure}, we consider three ways to measure consistency, focusing on different characteristics of the distribution: \textbf{agreement-based}, as mentioned before; 
\textbf{entropy-based}, which is based on the normalized entropy of the generation distribution; and \textbf{first-second-distance-based} (FSD), which measures the percentage difference in samples agreeing with the majority and second-majority answers.

We study the effectiveness of each consistency metric when applied to confidence calibration on both open-source (LLAMA, Mistral) and closed-source LLMs (Codex, GPT-3.5-turbo, GPT-4), 
and on nine datasets of four diverse reasoning tasks (Math Reasoning, Multi-Hop QA, Planning, Relational Reasoning).
Our experiments reveal several interesting findings:  (i) On average, all three consistency metrics significantly outperform existing post-hoc calibration baselines such as probabilistic and verbalized confidence extraction methods. 
(ii) When prompted to generate explanations before the answer, LMs exhibit markedly improved calibration. 
(iii) Scaling model size appears to enhance calibration, whereas instruction-tuning shows a negative effect. Increasing the number of generation samples leads to more accurate calibration, with notable improvements observed even with as few as 3-5 samples. 
(iv) We show in an oracle case study that consistency not only offers more reliable confidence estimates, but also holds the potential to enhance model performance on end tasks.

Our contributions are as follows:\\
(a) We systematically study three approaches for confidence calibration through sample consistency, and validate their superiority compared to existing post-hoc calibration baselines.\\
(b) We provide a detailed analysis of factors influencing calibration properties of LLMs, revealing diverse insights.\\
(c) We provide researchers with a flow chart (Appendix~\ref{decision-tree}) to help them pick the most effective consistency measure based on the characteristics of their model.
\section{Related Work}


\minisection{Confidence Calibration in LMs.} Traditional calibration methods, such as probabilistic \cite{calibration-guo}, ensemble-based \cite{lakshminarayanan_simple_2017, gal_dropout_2016}, and density-based approaches \cite{lee_simple_2018, yoo_detection_2022}, have proved effective in better calibrating the confidence in white-box LMs. These methods require access to the model logits and/or their pretraining data, involve retraining multiple copies of the same model, or necessitate another dedicated calibration dataset. With the advent of LLMs, they become overly expensive and sometimes even inapplicable to closed-source LLMs. To this end, several post-hoc approaches have been developed. \citet{kadavath2022language} prompt the model to estimate the probability of its generated response being ``True'', while \citet{lin2022teaching} and \citet{mielke-etal-2022-reducing} investigate whether the model can directly verbalize its confidence (e.g., ``highly confident'', or ``80\% confident''). Another line of work focuses on calibrating confidence with sample consistency \cite[][i.a.]{selfconsistency, selfcheckgpt, xiong2023can, portillo-wightman-etal-2023-strength}, which only needs input and output access to the model. However, existing studies have only focused on agreement-based measures of consistency, resulting in potential overconfidence.
This necessitates a systematic study on how to best elicit confidence from consistency.

\minisection{Consistency.} The term ``Consistency'' has been used to refer to multiple concepts in NLP, including factual alignment \cite{tam2022evaluating}, logical soundness \cite{nye2021improving}, agreement within diverse outputs \cite{selfconsistency}, among others. 
We use the term ``consistency'' to refer to the uniformity in the distribution of multiple model generations, as measured by three metrics in Figure~\ref{fig:main_figure}.

\minisection{Reasoning Strategies in LLMs.}
LLMs exhibit impressive reasoning capabilities with in-context learning. Besides standard prompting \cite{brown_language_2020}, explanation-based prompting, where models produce a reasoning chain before the answer, brings a notable performance gain. The explanation can be in the form of free-text \cite{cot}, decomposed subquestions \cite{shridhar2022automatic, zhou2023leasttomost}, or structured symbolic language \cite{pot, fcot}. We study how calibration can be influenced by representative strategies from each category. 

\section{Method}

Consistency over multiple generations can be used as an indicator for understanding the confidence associated with model predictions. It has been studied in the past for logit-based uncertainty estimation such as model ensembling \cite{lakshminarayanan_simple_2017} and we extend it to multiple generations in LLMs. 
For a given input $x$, we generate a set of $n$ candidate outputs $\hat{s}_1, \dots \hat{s}_n$.
From each sample $\hat{s}_i$, we parse the final answer $\hat{a_i}$ using regex matching. We do a majority voting over the entire answer (multi-)set $\mathbf{a} = \{ \hat{a_1} \dots \hat{a_n} \}$ to get the most-voted answer $\Bar{a} = \argmax_a \sum \nolimits_{i=1}^n \mathbbm{1} (\hat{a_i} = a)$, where $a$ takes on values from the set of unique answers $\mathbf{\Bar{a}}$.

\subsection{Calibration with Consistency}
This section presents three ways to measure consistency: agreement-based, entropy-based, and first-second-distance-based (FSD).
From each measure, we aim to obtain a confidence score $\texttt{conf}(x, \Bar{a})$ for each input $x$ to calibrate the correctness of the prediction.

\paragraph{Agreement-based.} 
Following previous work \cite{selfconsistency, xiong2023can}, we compute the agreement-based consistency by calculating the percentage of answers in $\mathbf{a}$ that agree with the most-voted answer $\Bar{a}$. In other words, agreement-based consistency, \texttt{Agree}($\Bar{a}$) is defined as:
\begin{equation}
   \texttt{Agree}(\Bar{a}) =  \frac{1}{n}\sum \limits_{i=1}^n \mathbbm{1} (\hat{a_i} = \Bar{a})
\end{equation}

\paragraph{Entropy-based.} 
In classification tasks, the entropy of output class probabilities has been used to estimate prediction uncertainty \cite{Gal2016UncertaintyID}. 
We extend this idea to the distribution of multiple model generations to understand the uncertainty in solving an open-ended reasoning problem, where a lower entropy indicates a more consistent distribution.

To calculate entropy-based consistency, we first obtain a set of answers without duplicates  $\mathbf{\Bar{a}}$. 
Then, we define entropy-based consistency, \texttt{Ent}($\mathbf{a}$) as:
\begin{equation}
    \texttt{Ent}(\mathbf{a}) = 1 - (- \frac{1}{\log(|\mathbf{\Bar{a}}|)} \sum \limits_{i=1}^{|\mathbf{\Bar{a}}|}\ p_i\ \log(p_i))
\end{equation}

\noindent where, the cardinality of the unique answer set $|\mathbf{\Bar{a}}|$ denotes the number of unique answers in the set $\mathbf{a}$ and the probability $p_i$ is the normalized frequency of each unique answer $\Bar{a_i}$ in the multi-set $\mathbf{a}$.

Note that the normalized entropy on the right side of the equation is subtracted from $1$ to reverse the range between $[0,1]$ as the lower the entropy, the more consistent the samples are, and thereby the higher the elicited confidence is. 

\begin{figure*}[t!]
\centering
\includegraphics[width=0.95\textwidth]{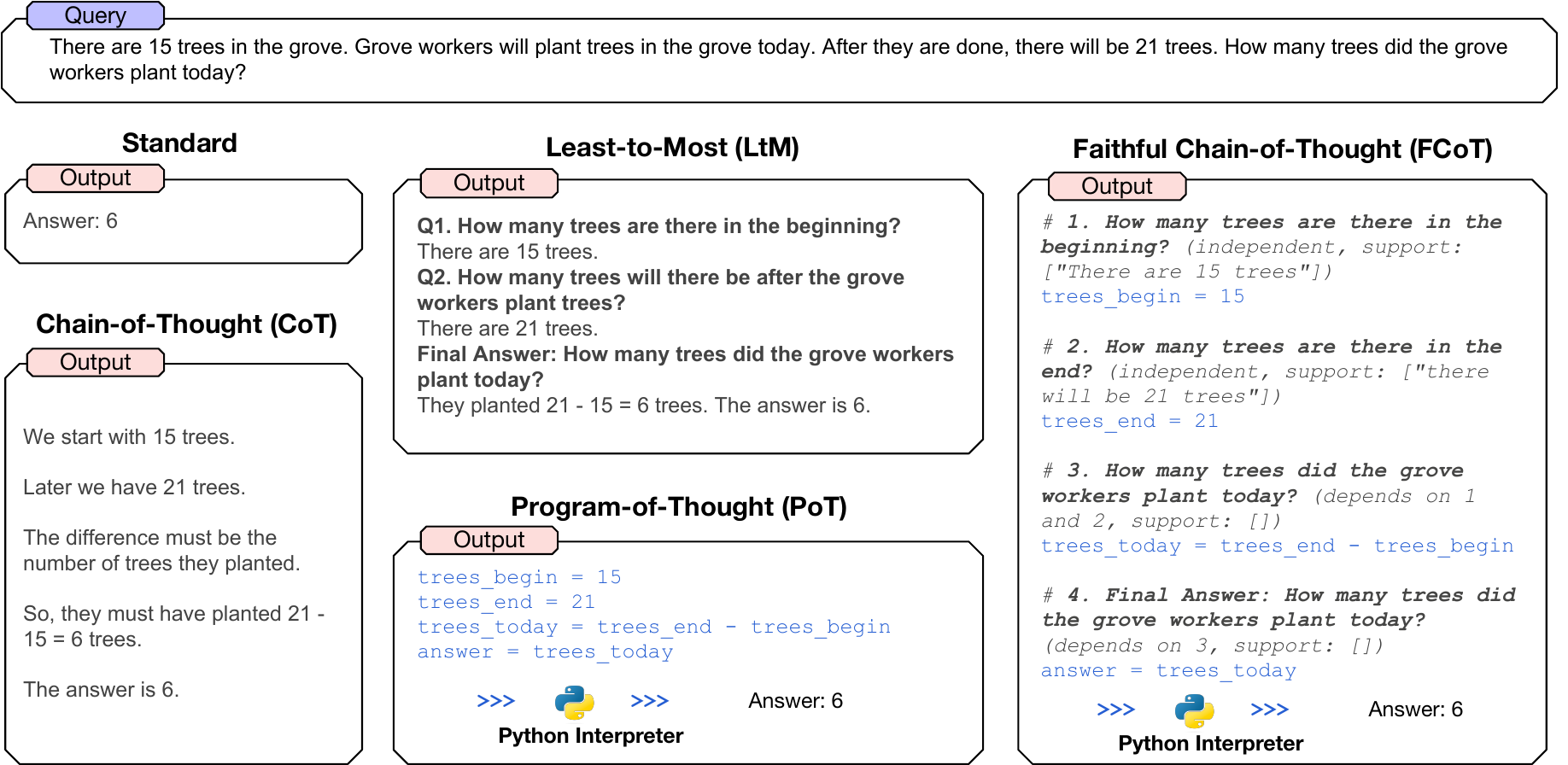}
\caption{We study how prompting strategies affect confidence calibration. Here is an example of a math question using the five prompting strategies that we consider.}
\vspace{-0.13in}
\label{fig:prompting_strategies}
\end{figure*}

\paragraph{FSD-based.} Since the entropy-based measure considers all unique answers that might be skewed toward the tail of the frequency distribution, and agreement-based consistency relies on the most-voted answer, we propose a third alternative, FSD.
To compute FSD-based consistency, we consider the top two most-voted answers ($\Bar{a}$ and $\Bar{\Bar{a}}$) and calculate the corresponding agreements $\texttt{Agree}(\Bar{a})$ and $\texttt{Agree}(\Bar{\Bar{a}})$.
Then, we use the difference between the two to compute the FSD-based consistency, \texttt{FSD}($\mathbf{a}$):
\begin{equation}
   \texttt{FSD}(\mathbf{a}) = \texttt{Agree}(\Bar{a}) - \texttt{Agree}(\Bar{\Bar{a}})
\end{equation}

This metric is particularly useful for cases when the model is unsure about the most-voted answer and places high confidence in the top two predictions. In such cases, an FSD-based consistency measure can avoid overconfidence based on the most-voted answer alone.  


\section{Experimental Setup}
\label{sec:implemetation_details}

\paragraph{Baselines.} We compare consistency-based calibration with four post-hoc methods:\footnote{See Appendix~\ref{appendix:baselines} for sample prompts.}

\noindent\textbullet\ \textbf{Raw logits} (\textbf{logit}) directly considers the probability of the generation as the confidence. Specifically, we take the exponential of the average log probability of all tokens in the output sequence, which is equivalent to the reciprocal of perplexity.

\noindent\textbullet\ \textbf{P(True)} \cite{kadavath2022language} prompts the model to judge the truthfulness of its generation and considers the normalized probability assigned to the `True' token as its confidence. In our experiment, we consider both 0-shot and 8-shot prompting (\textbf{ptrue\textsubscript{0-shot}} and \textbf{ptrue\textsubscript{8-shot}}).

\noindent\textbullet\ \textbf{Verbalized Confidence} \cite{lin2022teaching} prompts the model to explicitly verbalize its confidence in its generation as a linguistic expression (\textbf{verb\textsubscript{ling}}) from ``almost no chance'',  ``likely'', ...,  to ``almost certain'', which are mapped to a confidence level; or a percentage (\textbf{verb\textsubscript{percent}}) from 0 to 100, directly used as the confidence score.

We compare consistency-based calibration with only verbalized methods for GPT-3.5-turbo and GPT-4 since probabilities are not accessible, and with only logit for open-source models due to high computation cost (see details in Appendix~\ref{appendix:implementation_details_open}).

\paragraph{Tasks.} We experiment with nine datasets from four reasoning tasks following 
\citet{fcot}:\footnote{See Appendix~\ref{appendix:dataset_details} for dataset statistics and examples.}

\noindent\textbullet\ \textbf{Math Word Problems (MWPs)}: ASDiv \cite{miao_diverse_2020}, GSM8K \cite{cobbe_training_2021}, MultiArith \cite{roy_solving_2015}, and SVAMP \cite{patel_are_2021}.

\noindent\textbullet\ \textbf{Multi-hop QA}: StrategyQA \cite{geva_did_2021}, and two BIG-BENCH datasets \cite{big-bench_collaboration_beyond_2021}, Date Understanding and Sports Understanding.

\noindent\textbullet\ \textbf{Planning}: SayCan \cite{ahn_as_2022}.

\noindent\textbullet\ \textbf{Relational inference}: CLUTRR \cite{sinha_clutrr_2019}.

\paragraph{Evaluation metrics.} We use two established calibration metrics \cite{geng_survey_2023}. Let $\mathcal{D} = \{(x_j, y_j)\}, j \in \{1, \dots, N\}$ be the evaluation set used to measure calibration. Here $x_j$'s are inputs and $y_j$'s are ground-truth answers.

\noindent\textbullet\ \textbf{Brier Score} \cite{brier1950verification} measures the mean squared error between the confidence and the prediction correctness:
\begin{equation}
    BS = \frac{1}{N} \sum_{i=1}^N (\texttt{conf}(x_j, \hat{y}_j) - \mathbb{I}(\hat{y}_j = y_j) )^2
\end{equation}  
where the indicator $\mathbb{I}(\cdot)$ equals 1 when the prediction is correct, and otherwise it is 0. 

\noindent\textbullet\ \textbf{Expected Calibration Error (ECE)} \cite{calibration-guo} partitions the confidence scores $\{\texttt{conf}(x_j, \hat{y}_j)\}$ into $M$ equally spaced buckets $\{B_m\}_{m=1}^M$, with $B_m$ containing samples with confidence within the interval $(\frac{m-1}{M}, \frac{m}{M}]$. ECE is then defined as:
\begin{equation}
\label{eq:ece}
   ECE= \sum_{m=1}^M \frac{\lvert B_m \rvert}{N} \lvert \texttt{acc}(B_m) - \texttt{conf}(B_m)\rvert
\end{equation}
where the averaged accuracy and confidence in each bin $B_m$ are defined as: 
\begin{align}
\texttt{acc}(B_m) &= \frac{1}{\lvert B_m \rvert} \sum_{x_j \in B_m} \mathbb{I}(\hat{y}_j = y_j)\\
\texttt{conf}(B_m) &= \frac{1}{\lvert B_m\rvert} \sum_{x_j \in B_m} \texttt{conf}(x_j, \hat{y}_j)  
\end{align}
Since ECE has known issues such as sensitivity to the bin size \cite{geng_survey_2023}, we use the Brier Score as the main metric and put ECE results in the Appendix.

\begin{table*}[!ht]
    \centering
    \scalebox{0.75}{
    \begin{tabular}{p{3cm}|>{\raggedleft\arraybackslash}p{1.8cm}>{\raggedleft\arraybackslash}p{1.8cm}>{\raggedleft\arraybackslash}p{1.8cm}|>{\raggedleft\arraybackslash}p{1.8cm}>{\raggedleft\arraybackslash}p{1.8cm}>{\raggedleft\arraybackslash}p{1.8cm}>{\raggedleft\arraybackslash}p{1.8cm}>{\raggedleft\arraybackslash}p{1.8cm}}
    \toprule
        \textbf{LM} & \multicolumn{3}{c|}{\textbf{Consistency Metrics}} & \multicolumn{5}{c}{\textbf{Baselines}} \\

         & \textbf{entropy} & \textbf{agreement} & \textbf{FSD} & \textbf{verb\textsubscript{ling}} & \textbf{verb\textsubscript{percent}} & \textbf{logit} & \textbf{ptrue\textsubscript{0-shot}} & \textbf{ptrue\textsubscript{8-shot}} \\ \midrule
        Codex & .175\phantom{†} & \textbf{.151}† & \underline{.159}† & .249 & .249 & .209 & .188 & .179 \\ 
        GPT-3.5-turbo & \textbf{.205}† & .221† & \underline{.207}† & .271 & .273 & n/a & n/a & n/a \\
        GPT-4 & \underline{.116}† & .119† & \textbf{.114}† & .154 & .181 & n/a & n/a & n/a \\ 
        \bottomrule
    \end{tabular}
    }
    \caption{Consistency metrics result in better Brier Scores than baselines ($\downarrow$) for closed-source models. Scores are averaged across four domains and five prompting strategies. The best scores are \textbf{in bold} and the second-best scores are \underline{underlined}. † indicates that the consistency metric performs statistically significantly better than the best-performing baseline ($p<0.05$).}
    \label{table:results_5.1_close_avg_only}
\end{table*}
\begin{table}[!ht]
    \centering
        \addtolength{\tabcolsep}{-3pt}
    \scalebox{0.7}{
    \begin{tabular}{p{3cm}|>{\raggedleft\arraybackslash}p{1.5cm}>{\raggedleft\arraybackslash}p{1.5cm}>{\raggedleft\arraybackslash}p{1.5cm}|>{\raggedleft\arraybackslash}p{1.5cm}}
    \toprule
        \textbf{LM} & \multicolumn{3}{c|}{\textbf{Consistency Metrics}} & \multicolumn{1}{c}{\textbf{Baselines}} \\

        & \textbf{entropy} & \textbf{agree} & \textbf{FSD} & \textbf{logit} \\ 
        \midrule
         LLaMA-7B & .241† & \textbf{.232}† & \underline{.235}† & .474 \\ 
        
         LLaMA-13B & .222† & \textbf{.204}† & \underline{.211}† & .389 \\ 
        
         LLaMA-70B & .182† & \textbf{.154}† & \underline{.165}† & .252 \\ 
        
         Mistral-7B & .205† & \textbf{.183}† & \underline{.191}† & .324 \\ 
        
         Mistral-7B-instruct & .220† & \underline{.216}† & \textbf{.215}† & .384 \\
        \bottomrule
    \end{tabular}
    }
    \vspace{-0.03in}
    \caption{Consistency metrics result in better Brier Scores ($\downarrow$) than the logit baseline for open-source models.}
    \vspace{-0.13in}
    \label{table:results_5.1_open_avg_only}
\end{table}

\paragraph{Prompting strategies.} We compare five prompting strategies in Figure~\ref{fig:prompting_strategies}: \textbf{standard} prompting, where an exemplar contains only the query and the answer; \textbf{CoT} \cite{cot}, which additionally includes a Natural Language (NL) reasoning chain; \textbf{Least-to-Most (LtM)} \cite{zhou2023leasttomost}, which decomposes the question into NL subquestions; \textbf{Program of Thoughts (PoT)}\footnote{Also called Program-Aided Language Model (PAL) in the concurrent work by \citet{gao_pal_2023}.} \cite{chen_close_2023}, which solves the query in Symbolic Language (SL); and \textbf{Faithful CoT (FCoT)} \cite{fcot}, which interleaves NL subquestions and SL solutions. We use the same prompts from \citet{fcot}, with the same number of shots for each strategy (6 to 10, depending on the dataset).  

\paragraph{LMs.} We consider the following LLMs: LLaMA (7B/13B/70B), Mistral (7B/7B-instruct), Codex, GPT-3.5-turbo, and GPT-4.\footnote{See checkpoint names and computational resources in Appendix~\ref{appendix:implementation_details}.}

\paragraph{Sampling Strategy.} In our main experiments, we sample $n=40$ candidate outputs with a temperature of $T=0.4$ for each input following \citet{fcot}, and analyze other values of $n$ and $T$ in Section~\ref{sec:analysis}. We select the majority-voted answer as the final answer, following \citet{selfconsistency}.

\section{Results}
\label{sec:results}

We study our research question -- \emph{how can we best elicit a model's confidence from the consistency of multiple generations?} -- from two perspectives: which \textbf{calibration method} is the most effective, and how does the \textbf{prompting strategy} affect a model's calibration properties?

\subsection{Comparing Calibration Methods}

\begin{figure*}[t!]
\centering
\includegraphics[width=0.9\textwidth]{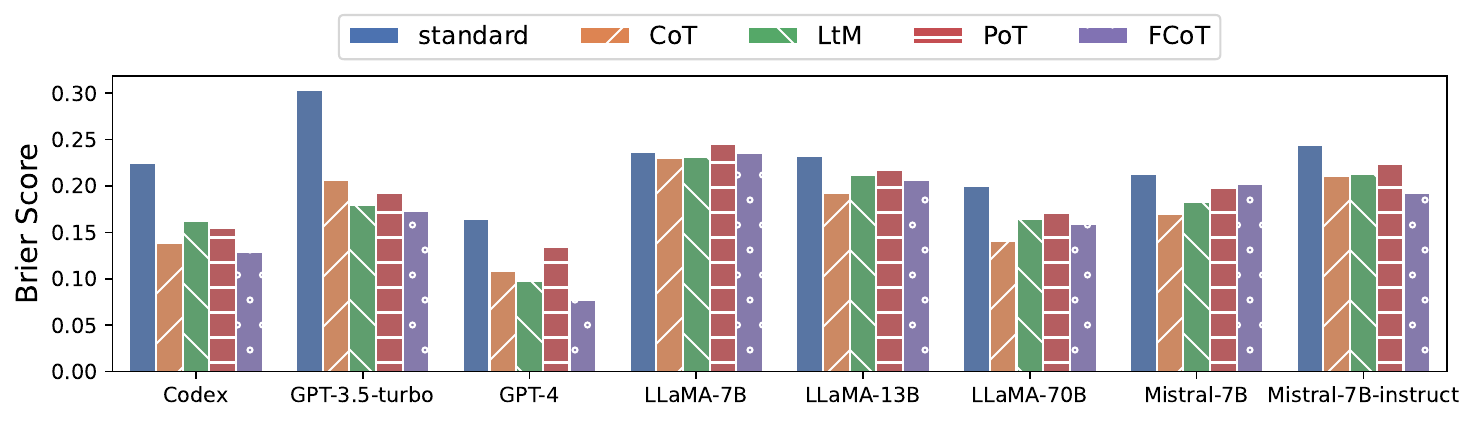}
\caption{Brier Scores ($\downarrow$) are improved with explanation-based prompting strategies, with Chain of Thought (CoT) and Faithful CoT (FCoT) performing the best. Scores here are averaged across all datasets and consistency metrics.}
\vspace{-0.13in}
\label{fig:results_compare_prompts}
\end{figure*}

We compare all calibration methods in \autoref{table:results_5.1_close_avg_only} and \autoref{table:results_5.1_open_avg_only}, which show the average Brier Score for closed-source and open-source LMs averaged across all datasets. See full results in Appendix~\ref{appendix:calib_results_all_datasets}.

\paragraph{Consistency-based methods are more effective than baselines.} Our results suggest a clear advantage of consistency-based calibration methods over the baselines. Averaging across domains, all three consistency metrics almost always result in a significantly lower Brier Score ($p<0.05$) than the best-performing baseline. This trend also holds across the vast majority of the LMs and domains tested. In rare exceptions in the Relational Inference and Planning domains, the optimal consistency metric often performs statistically the same as the baseline.

\paragraph{Agreement-based consistency works best for open-source models and Codex, while FSD and entropy for the other closed-source models.} Among all three consistency metrics, which one is the most effective? We compare the statistical significance between the performance differences of the three metrics in Table~\ref{table:3_consistency_significance} in Appendix~\ref{appendix:additional_results}. For closed-source models, agreement is the most effective metric for Codex ($p < 0.05$), while entropy and FSD are closely competing within a negligible performance gap ($\delta_{BS} \leq$ 0.002, $p\geq 0.05$) for GPT-3.5-turbo and GPT-4. Meanwhile, open-source models predominantly favor agreement ($p < 0.05$), with FSD closely following as the second-best metric. The sole exception to this trend is in the case of Mistral-7B-instruct, where FSD leads over agreement by a slim margin (0.215 vs. 0.216, $p \geq 0.05$). 

When dissecting the results domain-wise, entropy consistently emerges as the favored metric in Relational Inference across all tested models, whereas the Planning domain shows a predominant preference for agreement for all but one model (GPT-3.5-turbo).

Synthesizing these findings, agreement is the most effective consistency metric for Codex and most open-source models, closely followed by FSD. For GPT-3.5-turbo and GPT-4, FSD and entropy are closely matched in effectiveness. A conjectured reason for this discrepancy could be the lack of Reinforcement Learning from Human Feedback (RLHF) in Codex and open-source models, unlike GPT-3.5-turbo and GPT-4. However, the exact cause remains indeterminate due to the unavailability of a minimal pair of models with and without RLHF for a controlled comparison.

\paragraph{Takeaways.} Our findings indicate that consistency metrics offer a more reliable measure of confidence than baselines. Among all consistency metrics, FSD stands out as a robust default selection, maintaining stable performance across various models and domains, often achieving the highest or near-highest performance.  

\begin{figure*}[t!]
\centering
\includegraphics[width=0.8\textwidth]{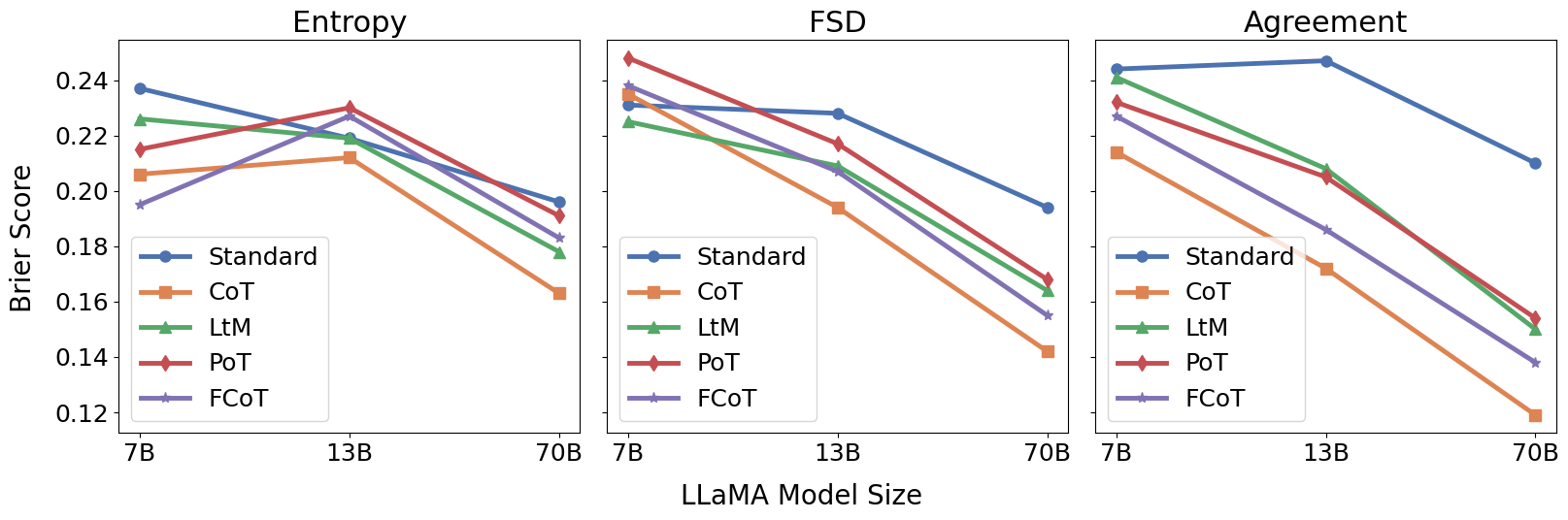}
\caption{The Brier Score ($\downarrow$) tends to improve as the model size increases for the 3 studied calibration metrics across most of the prompting techniques we consider.}
\vspace{-0.13in}
\label{fig:scaling-all}
\end{figure*}

\subsection{The Role of Explanations}
Does the prompting strategy influence how well a model can be calibrated? Here, we compare \textbf{standard} prompting, where the model only predicts the answer, against four \textbf{explanation-based} prompting strategies (CoT, LtM, PoT, and FCoT), where the model produces a reasoning chain before the answer. Figure~\ref{fig:results_compare_prompts} shows the results for each prompting strategy averaged across consistency metrics.

\paragraph{Explanations make LMs better-calibrated.} When models are prompted to generate any form of explanation (CoT, LtM, PoT, and FCoT) before the answer, they exhibit a marked improvement in calibration error ($p<0.05$). This finding holds across the board with the only exception of LLAMA-7B, which appears to be indifferent to the prompting strategy. The benefit of explanations on calibration is especially evident in larger models, mirroring the observed correlation between accuracy and model size with explanations \cite{cot}.

\paragraph{GPT models are best calibrated with FCoT, while most open-source models are best calibrated with CoT.} The calibration efficacy of GPT models (Codex, GPT-3.5-turbo, GPT-4) and Mistral-7B-instruct is maximized through FCoT prompting, which interleaves NL and SL. Conversely, when it comes to open-source models, CoT in pure NL appears to be the most effective in enhancing calibration. This contrast underscores a potential difference in how these closed-source and open-source models process and benefit from prompts involving explanations.

\paragraph{Takeaways.} Including explanations in prompts not only bolsters LMs' performance (Figure~\ref{table:accuracy}) but also makes them better-calibrated. This dual benefit suggests that the process of generating explanations potentially aids models in better processing and reasoning about the tasks at hand, leading to outputs more closely aligned with expectations. 
\section{Analysis}
\label{sec:analysis}

In this section, we examine how scaling, instruction-tuning, and sample size affect the calibration properties across various LMs. 

\subsection{How Does Scaling Affect Calibration?}

We study how an increase in model parameters impacts different consistency
metrics. \autoref{fig:scaling-all} compares Brier Score across all reasoning strategies (standard, CoT, LtM, PoT, and FCoT) for all three consistency metrics (Entropy, Agreement, and FSD) for different sized LLaMA models (7B, 13B, and 70B), in order to understand the effect of scaling on calibration. 
We observe that the average Brier Score across datasets goes down for all consistency metrics as the model scales up; suggesting that \emph{scaling supports calibration}. In other words, the larger the model, the better it is calibrated across the various tasks studied in this paper. 

Moreover, we observe that for LLaMA-7B, all prompting strategies have a very similar Brier Score (as seen from the left side in the \autoref{fig:scaling-all}). As the model scales up to 70B, the gap increases (to the right of \autoref{fig:scaling-all}), especially between standard prompting and explanation-based strategies (all others). This shows that \emph{explanation improves calibration with scale}. 

\subsection{How Does Instruction-Tuning Affect Calibration?}
\begin{figure}[t!]
\centering
\includegraphics[width=\columnwidth]{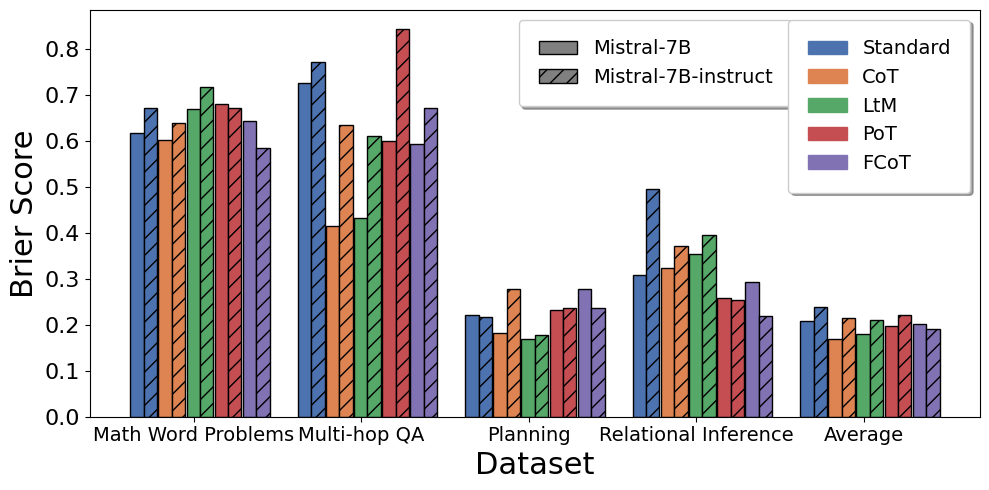}
\caption{Surprisingly, the non-instruction-tuned model (Mistral-7B) has better Brier Scores ($\downarrow$) compared to an instruction-tuned model (Mistral-7B-instruct) across nearly all of our prompting strategies and tasks.}
\vspace{-0.13in}
\label{fig:instruction-tuned}
\end{figure}

\begin{figure*}[t!]
\centering
\includegraphics[width=0.95\textwidth]{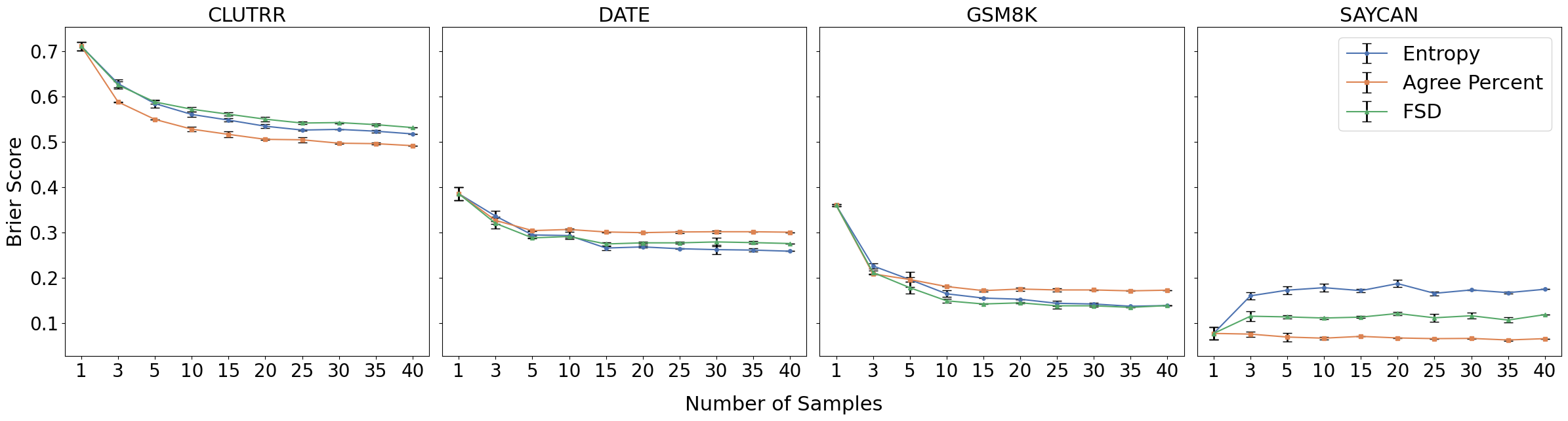}
\vspace{-0.1in}
\caption{Brier Scores ($\downarrow$) improve as we increase the number of samples for 3 of the 4 datasets. Results are obtained with GPT-3.5-turbo and CoT prompting.}
\vspace{-0.1in}
\label{fig:effect_of_n}
\end{figure*}

\begin{figure*}[t!]
\centering
\includegraphics[width=0.95\textwidth]{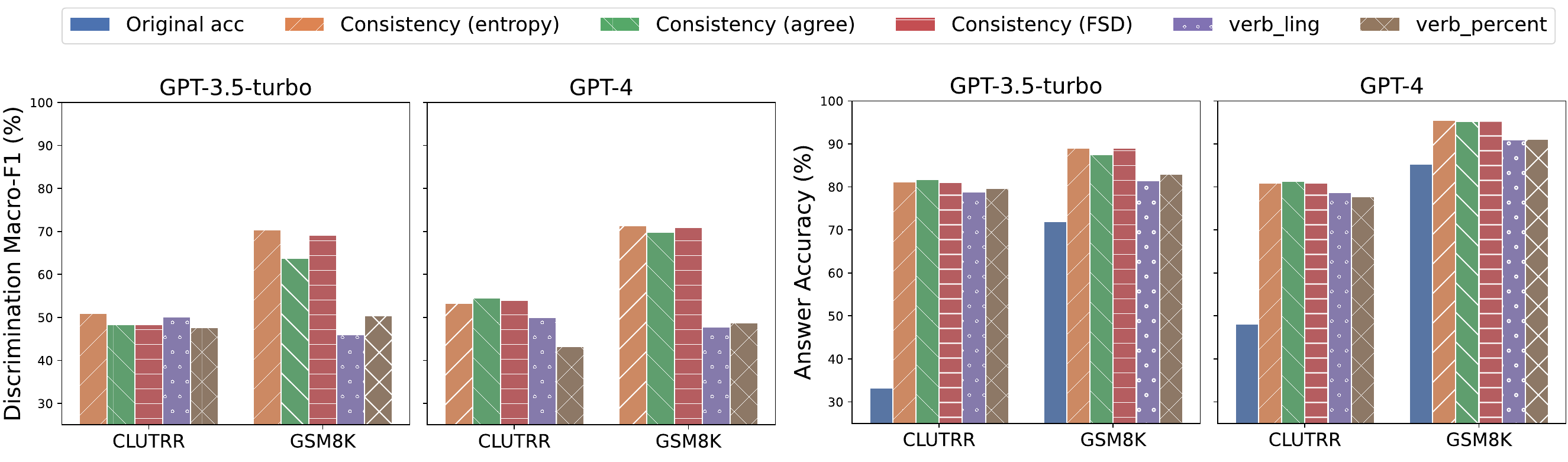}
\caption{Left: Consistency-based calibration outperforms the verbalized baselines in their ability to discriminate the correctness of predictions measured by Macro-F1 ($\uparrow$). Right: Consistency-based calibration leads to a larger improvement in answer accuracy ($\uparrow$) after correcting the top-k\% most uncertain predictions with oracle answers. Scores in these figures are averaged across all prompting strategies.}
\vspace{-0.13in}
\label{fig:selfcorrect}
\end{figure*}


To understand the effect of instruction-tuning, we compare the calibration properties of Mistral-7B and Mistral-7B-instruct across the four tasks we studied. \autoref{fig:instruction-tuned} demonstrates that in general \emph{instruction-tuning leads to worse calibration properties}, which is analogous to findings from the past works of \citet{kadavath2022language}. However,
\emph{faithful prompting strategies improve calibration for instruction-tuned models}
 as shown by the lower Brier Scores for FCoT on almost all datasets and the final average in \autoref{fig:instruction-tuned}.

\subsection{How does the Number of Generated Outputs impact Calibration?}

We analyze the usefulness of consistency-based calibration by generating different numbers of output samples and calculating different consistency metrics over them. \autoref{fig:effect_of_n} demonstrates that generating \emph{more samples can lead to better calibration scores}, as indicated by a downward trend in the Brier Score. We observe the improvement in the Brier Score as a function of the number of samples and the decision of the appropriate number of samples can be made based on the available computational budget and the desired calibration properties. Brier Scores usually saturate after $15-20$ samples, with a sharp drop at the beginning. For budget constraints, $3-5$ samples is a good choice.

\section{Case Study: Does Calibration Help Improve Model Performance?}

Beyond calibrating trust in model predictions, can consistency metrics contribute to improving task performance? To explore this, we perform a case study with GPT-3.5-turbo and GPT-4 on GSM8K and CLUTRR datasets from the MWP and Relation Reasoning domains respectively. We compare the consistency metrics against other calibration baselines in two experiments: \textbf{discriminating prediction correctness} and \textbf{improving final answer accuracy}. 

In the first experiment, given a model's predictions $\hat{Y}$ on a dataset $X$, our goal is to differentiate the correctness of each prediction $\hat{y_i}$ with the confidence $\texttt{conf}(x_i, \hat{y}_i)$ provided by any calibration method. Identifying incorrect predictions is the first step for performance improvement, and it can be integrated into any self-correction pipeline \cite[][i.a.]{self-refine, shridhar2023screws}. To test discrimination efficacy, we tune an optimal threshold $\theta$ for each calibration method on a development set.\footnote{See Appendix~\ref{appendix:implementation_details} for tuning details.} If the provided confidence score $\texttt{conf}(x_i, \hat{y}_i)$ is above $\theta$, we consider the model prediction as correct, otherwise incorrect. Then, we evaluate the discrimination performance of each calibration method on the test set. 
The results, illustrated in Figure~\ref{fig:selfcorrect} (left), indicate that consistency metrics significantly outstrip verbalized baselines in discriminating correct and incorrect predictions, with the effect being most pronounced on the GSM8K dataset (more than doubled Macro-F1).
All three consistency metrics share this trend, except for the only case of GPT-3.5-turbo on CLUTRR, where entropy outperforms the optimal baseline, yet the gap between all methods is small. \looseness=-1

The second experiment assesses the impact of a calibration method on answer accuracy, assuming an oracle subsequent self-correction mechanism. Given a model's predictions $\hat{Y}$ on a dataset $X$, we identify the top-k\% most uncertain predictions, $\hat{Y_{-}}$, which are those with the lowest confidence scores according to the calibration method, as incorrect. 
This fixed $k$ is chosen to be the true error rate of all model predictions, i.e., $k = 1-\text{acc}(\hat{Y}, X)$. Finally, we correct $\hat{Y_{-}}$ with the ground-truth answers and evaluate the resulting accuracy. 
As shown in Figure~\ref{fig:selfcorrect} (right), post-correction accuracy exceeds original accuracy to the greatest extent when applying consistency-based calibration.

In both experiments, entropy and FSD are equally effective on GSM8K for both models, while agreement and entropy lead on CLUTRR for each model. In summary, consistency provides not just a measure of prediction trust, but can also contribute to enhanced model performance assuming ideal self-correction mechanisms. 

\section{Conclusion}


We investigate the effectiveness of eliciting confidence in LLMs from sample consistency, using entropy and FSD as extensions of the naive agreement-based consistency measure. Through extensive evaluations on various open- and closed-source models and nine reasoning datasets, we demonstrate the superiority of these methods over traditional post-hoc verbalized and probabilistic calibration techniques. Further analysis shows that explanation generation, model scaling, and larger sample sizes improve calibration, while instruction-tuning has a counter-effect. In addition to providing more reliable confidence estimates, consistency measures also contribute to improved model performance when integrated into an ideal self-correction pipeline. Finally, our work provides practical guidance for selecting the most appropriate consistency metrics for calibration based on different model types, sizes, and inference tasks, paving the way for more reliable and trustworthy applications of LLMs in various domains (a prescriptive starting guide is provided in \autoref{decision-tree}). 
\section*{Limitations}

We acknowledge several limitations in this study. First, no single consistency metric emerged as universally superior across all LMs and datasets. To this end, we provide recommendations for context-specific metric selection. Second, we choose the sample size as $n=40$ in our main experiments following the default setting in \citet{wang2023survey}, which entails a considerable cost. Nevertheless, it is not necessary to use such a large sample size in practice, since we find that the calibration performance already sees a notable improvement with 3 to 5 generations, and saturates around 15 to 20 generations. Third, we have only used the temperature value of $T=0.4$ following previous work in our experiments. An analysis on the effect of different temperature values will shed light on the robustness of consistency-based calibration. Fourth, our approach only focuses on measuring the consistency among \textit{final answers}, overlooking intermediate steps in various prompting techniques. Future work can explore how to calibrate the model confidence in each intermediate step in a reasoning chain. Finally, the deprecation of Codex as of Jan 4 2024 poses a challenge for replicating some of our results. Despite this, we have preserved all model outputs to ensure reproducibility as far as possible.

\section*{Ethical Considerations}

Our work in the area of calibration explores the importance of trustworthiness and transparency in LLMs. However, our analysis was performed on the pre-trained models and in many cases it might be biased towards a specific task/domain. Given the nature of these LLMs, the biases in the dataset, training, and post-processing may affect the decision making capabilities of such systems and must not be used in high-stakes scenarios despite the positive calibration results. We would also like to mention that we have no control over the alignment of these LLMs during instruction-tuning, and we cannot comment on whether the alignment was done with ethical standards and societal norms in mind. 

We also believe that the proprietary nature of many state-of-the-art LLMs studied in this work requires a lot of computation and incurs high API costs. This hinders the accessibility of such models to a wider research community. However, we have studied smaller models such as LLaMA and Mistral in this work to make our work accessible to communities with limited resources. 

\section{Acknowledgements}
This research is based upon work supported in part by the Air Force Research Laboratory (contract FA8750-23-C-0507), the DARPA KAIROS Program (contract FA8750-19-2-1004), the IARPA HIATUS Program (contract 2022-22072200005), the NSF (Award 1928631), the Swiss National Science Foundation (Project No. 197155), and a Responsible AI grant by the Haslerstiftung. Approved for Public Release, Distribution Unlimited. The views and conclusions contained herein are those of the authors and should not be interpreted as necessarily representing the official policies, either expressed or implied, of AFRL, DARPA, IARPA, NSF, SNSF or the U.S. Government. 

We appreciate the support from OpenAI by extending our access to Codex through the Researcher Access Program. We also thank Xinran Zhao, Hongming Zhang, Jie Huang, Hangfeng He, Charles Yang, Lyle Ungar, Shehzaad Dhuliawala, and Dan Roth for their valuable feedback.

\bibliography{custom}

\begin{thebibliography}{42}
\expandafter\ifx\csname natexlab\endcsname\relax\def\natexlab#1{#1}\fi

\bibitem[{Ahn et~al.(2022)Ahn, Brohan, Brown, Chebotar, Cortes, David, Finn, Fu, Gopalakrishnan, Hausman, Herzog, Ho, Hsu, Ibarz, Ichter, Irpan, Jang, Ruano, Jeffrey, Jesmonth, Joshi, Julian, Kalashnikov, Kuang, Lee, Levine, Lu, Luu, Parada, Pastor, Quiambao, Rao, Rettinghouse, Reyes, Sermanet, Sievers, Tan, Toshev, Vanhoucke, Xia, Xiao, Xu, Xu, Yan, and Zeng}]{ahn_as_2022}
Michael Ahn, Anthony Brohan, Noah Brown, Yevgen Chebotar, Omar Cortes, Byron David, Chelsea Finn, Chuyuan Fu, Keerthana Gopalakrishnan, Karol Hausman, Alex Herzog, Daniel Ho, Jasmine Hsu, Julian Ibarz, Brian Ichter, Alex Irpan, Eric Jang, Rosario~Jauregui Ruano, Kyle Jeffrey, Sally Jesmonth, Nikhil~J. Joshi, Ryan Julian, Dmitry Kalashnikov, Yuheng Kuang, Kuang-Huei Lee, Sergey Levine, Yao Lu, Linda Luu, Carolina Parada, Peter Pastor, Jornell Quiambao, Kanishka Rao, Jarek Rettinghouse, Diego Reyes, Pierre Sermanet, Nicolas Sievers, Clayton Tan, Alexander Toshev, Vincent Vanhoucke, Fei Xia, Ted Xiao, Peng Xu, Sichun Xu, Mengyuan Yan, and Andy Zeng. 2022.
\newblock \href {https://arxiv.org/abs/2204.01691} {Do {As} {I} {Can}, {Not} {As} {I} {Say}: {Grounding} {Language} in {Robotic} {Affordances}}.

\bibitem[{{BIG-Bench collaboration}(2021)}]{big-bench_collaboration_beyond_2021}
{BIG-Bench collaboration}. 2021.
\newblock \href {https://github.com/google/BIG-bench/} {Beyond the {Imitation} {Game}: {Measuring} and extrapolating the capabilities of language models}.

\bibitem[{Brier(1950)}]{brier1950verification}
Glenn~W Brier. 1950.
\newblock Verification of forecasts expressed in terms of probability.
\newblock \emph{Monthly weather review}, 78(1):1--3.

\bibitem[{Brown et~al.(2020)Brown, Mann, Ryder, Subbiah, Kaplan, Dhariwal, Neelakantan, Shyam, Sastry, Askell, Agarwal, Herbert{-}Voss, Krueger, Henighan, Child, Ramesh, Ziegler, Wu, Winter, Hesse, Chen, Sigler, Litwin, Gray, Chess, Clark, Berner, McCandlish, Radford, Sutskever, and Amodei}]{brown_language_2020}
Tom~B. Brown, Benjamin Mann, Nick Ryder, Melanie Subbiah, Jared Kaplan, Prafulla Dhariwal, Arvind Neelakantan, Pranav Shyam, Girish Sastry, Amanda Askell, Sandhini Agarwal, Ariel Herbert{-}Voss, Gretchen Krueger, Tom Henighan, Rewon Child, Aditya Ramesh, Daniel~M. Ziegler, Jeffrey Wu, Clemens Winter, Christopher Hesse, Mark Chen, Eric Sigler, Mateusz Litwin, Scott Gray, Benjamin Chess, Jack Clark, Christopher Berner, Sam McCandlish, Alec Radford, Ilya Sutskever, and Dario Amodei. 2020.
\newblock \href {https://proceedings.neurips.cc/paper/2020/hash/1457c0d6bfcb4967418bfb8ac142f64a-Abstract.html} {Language models are few-shot learners}.
\newblock In \emph{Advances in Neural Information Processing Systems 33: Annual Conference on Neural Information Processing Systems 2020, NeurIPS 2020, December 6-12, 2020, virtual}.

\bibitem[{Chen et~al.(2021)Chen, Tworek, Jun, Yuan, Pinto, Kaplan, Edwards, Burda, Joseph, Brockman, Ray, Puri, Krueger, Petrov, Khlaaf, Sastry, Mishkin, Chan, Gray, Ryder, Pavlov, Power, Kaiser, Bavarian, Winter, Tillet, Such, Cummings, Plappert, Chantzis, Barnes, Herbert-Voss, Guss, Nichol, Paino, Tezak, Tang, Babuschkin, Balaji, Jain, Saunders, Hesse, Carr, Leike, Achiam, Misra, Morikawa, Radford, Knight, Brundage, Murati, Mayer, Welinder, McGrew, Amodei, McCandlish, Sutskever, and Zaremba}]{chen_evaluating_2021}
Mark Chen, Jerry Tworek, Heewoo Jun, Qiming Yuan, Henrique Ponde de~Oliveira Pinto, Jared Kaplan, Harri Edwards, Yuri Burda, Nicholas Joseph, Greg Brockman, Alex Ray, Raul Puri, Gretchen Krueger, Michael Petrov, Heidy Khlaaf, Girish Sastry, Pamela Mishkin, Brooke Chan, Scott Gray, Nick Ryder, Mikhail Pavlov, Alethea Power, Lukasz Kaiser, Mohammad Bavarian, Clemens Winter, Philippe Tillet, Felipe~Petroski Such, Dave Cummings, Matthias Plappert, Fotios Chantzis, Elizabeth Barnes, Ariel Herbert-Voss, William~Hebgen Guss, Alex Nichol, Alex Paino, Nikolas Tezak, Jie Tang, Igor Babuschkin, Suchir Balaji, Shantanu Jain, William Saunders, Christopher Hesse, Andrew~N. Carr, Jan Leike, Josh Achiam, Vedant Misra, Evan Morikawa, Alec Radford, Matthew Knight, Miles Brundage, Mira Murati, Katie Mayer, Peter Welinder, Bob McGrew, Dario Amodei, Sam McCandlish, Ilya Sutskever, and Wojciech Zaremba. 2021.
\newblock \href {https://arxiv.org/abs/2107.03374} {Evaluating {Large} {Language} {Models} {Trained} on {Code}}.

\bibitem[{Chen et~al.(2022)Chen, Ma, Wang, and Cohen}]{pot}
Wenhu Chen, Xueguang Ma, Xinyi Wang, and William~W Cohen. 2022.
\newblock \href {https://arxiv.org/abs/2211.12588} {Program of thoughts prompting: Disentangling computation from reasoning for numerical reasoning tasks}.
\newblock \emph{ArXiv preprint}, abs/2211.12588.

\bibitem[{Chen et~al.(2023)Chen, Yuan, Cui, Liu, and Ji}]{chen_close_2023}
Yangyi Chen, Lifan Yuan, Ganqu Cui, Zhiyuan Liu, and Heng Ji. 2023.
\newblock \href {https://doi.org/10.18653/v1/2023.acl-long.75} {A {Close} {Look} into the {Calibration} of {Pre}-trained {Language} {Models}}.
\newblock In \emph{Proceedings of the 61st {Annual} {Meeting} of the {Association} for {Computational} {Linguistics} ({Volume} 1: {Long} {Papers})}, pages 1343--1367, Toronto, Canada. Association for Computational Linguistics.

\bibitem[{Cobbe et~al.(2021)Cobbe, Kosaraju, Bavarian, Chen, Jun, Kaiser, Plappert, Tworek, Hilton, Nakano, Hesse, and Schulman}]{cobbe_training_2021}
Karl Cobbe, Vineet Kosaraju, Mohammad Bavarian, Mark Chen, Heewoo Jun, Lukasz Kaiser, Matthias Plappert, Jerry Tworek, Jacob Hilton, Reiichiro Nakano, Christopher Hesse, and John Schulman. 2021.
\newblock \href {https://arxiv.org/abs/2110.14168} {Training {Verifiers} to {Solve} {Math} {Word} {Problems}}.

\bibitem[{Gal(2016)}]{Gal2016UncertaintyID}
Yarin Gal. 2016.
\newblock \href {https://api.semanticscholar.org/CorpusID:86522127} {Uncertainty in deep learning}.

\bibitem[{Gal and Ghahramani(2016)}]{gal_dropout_2016}
Yarin Gal and Zoubin Ghahramani. 2016.
\newblock \href {http://proceedings.mlr.press/v48/gal16.html} {Dropout as a bayesian approximation: Representing model uncertainty in deep learning}.
\newblock In \emph{Proceedings of the 33nd International Conference on Machine Learning, {ICML} 2016, New York City, NY, USA, June 19-24, 2016}, volume~48 of \emph{{JMLR} Workshop and Conference Proceedings}, pages 1050--1059. JMLR.org.

\bibitem[{Gao et~al.(2023)Gao, Madaan, Zhou, Alon, Liu, Yang, Callan, and Neubig}]{gao_pal_2023}
Luyu Gao, Aman Madaan, Shuyan Zhou, Uri Alon, Pengfei Liu, Yiming Yang, Jamie Callan, and Graham Neubig. 2023.
\newblock \href {https://arxiv.org/abs/2211.10435} {Pal: Program-aided language models}.
\newblock In \emph{International Conference on Machine Learning}, pages 10764--10799. PMLR.

\bibitem[{Geng et~al.(2023)Geng, Cai, Wang, Koeppl, Nakov, and Gurevych}]{geng_survey_2023}
Jiahui Geng, Fengyu Cai, Yuxia Wang, Heinz Koeppl, Preslav Nakov, and Iryna Gurevych. 2023.
\newblock \href {https://arxiv.org/abs/2311.08298} {A {Survey} of {Language} {Model} {Confidence} {Estimation} and {Calibration}}.

\bibitem[{Geva et~al.(2021)Geva, Khashabi, Segal, Khot, Roth, and Berant}]{geva_did_2021}
Mor Geva, Daniel Khashabi, Elad Segal, Tushar Khot, Dan Roth, and Jonathan Berant. 2021.
\newblock \href {https://doi.org/10.1162/tacl_a_00370} {\textit{{Did} {Aristotle} {Use} a {Laptop}?} {A} {Question} {Answering} {Benchmark} with {Implicit} {Reasoning} {Strategies}}.
\newblock \emph{Transactions of the Association for Computational Linguistics}, 9:346--361.

\bibitem[{Guo et~al.(2017)Guo, Pleiss, Sun, and Weinberger}]{calibration-guo}
Chuan Guo, Geoff Pleiss, Yu~Sun, and Kilian~Q. Weinberger. 2017.
\newblock \href {http://proceedings.mlr.press/v70/guo17a.html} {On calibration of modern neural networks}.
\newblock In \emph{Proceedings of the 34th International Conference on Machine Learning, {ICML} 2017, Sydney, NSW, Australia, 6-11 August 2017}, volume~70 of \emph{Proceedings of Machine Learning Research}, pages 1321--1330. {PMLR}.

\bibitem[{Jiang et~al.(2023)Jiang, Sablayrolles, Mensch, Bamford, Chaplot, de~las Casas, Bressand, Lengyel, Lample, Saulnier, Lavaud, Lachaux, Stock, Scao, Lavril, Wang, Lacroix, and Sayed}]{jiang2023mistral}
Albert~Q. Jiang, Alexandre Sablayrolles, Arthur Mensch, Chris Bamford, Devendra~Singh Chaplot, Diego de~las Casas, Florian Bressand, Gianna Lengyel, Guillaume Lample, Lucile Saulnier, Lélio~Renard Lavaud, Marie-Anne Lachaux, Pierre Stock, Teven~Le Scao, Thibaut Lavril, Thomas Wang, Timothée Lacroix, and William~El Sayed. 2023.
\newblock \href {http://arxiv.org/abs/2310.06825} {Mistral 7b}.

\bibitem[{Jiang et~al.(2021)Jiang, Araki, Ding, and Neubig}]{jiang_how_2021}
Zhengbao Jiang, Jun Araki, Haibo Ding, and Graham Neubig. 2021.
\newblock \href {https://doi.org/10.1162/tacl_a_00407} {How can we know when language models know? on the calibration of language models for question answering}.
\newblock \emph{Transactions of the Association for Computational Linguistics}, 9:962--977.

\bibitem[{Kadavath et~al.(2022)Kadavath, Conerly, Askell, Henighan, Drain, Perez, Schiefer, Hatfield-Dodds, DasSarma, Tran-Johnson, Johnston, El-Showk, Jones, Elhage, Hume, Chen, Bai, Bowman, Fort, Ganguli, Hernandez, Jacobson, Kernion, Kravec, Lovitt, Ndousse, Olsson, Ringer, Amodei, Brown, Clark, Joseph, Mann, McCandlish, Olah, and Kaplan}]{kadavath2022language}
Saurav Kadavath, Tom Conerly, Amanda Askell, Tom Henighan, Dawn Drain, Ethan Perez, Nicholas Schiefer, Zac Hatfield-Dodds, Nova DasSarma, Eli Tran-Johnson, Scott Johnston, Sheer El-Showk, Andy Jones, Nelson Elhage, Tristan Hume, Anna Chen, Yuntao Bai, Sam Bowman, Stanislav Fort, Deep Ganguli, Danny Hernandez, Josh Jacobson, Jackson Kernion, Shauna Kravec, Liane Lovitt, Kamal Ndousse, Catherine Olsson, Sam Ringer, Dario Amodei, Tom Brown, Jack Clark, Nicholas Joseph, Ben Mann, Sam McCandlish, Chris Olah, and Jared Kaplan. 2022.
\newblock \href {http://arxiv.org/abs/2207.05221} {Language models (mostly) know what they know}.

\bibitem[{Lakshminarayanan et~al.(2017)Lakshminarayanan, Pritzel, and Blundell}]{lakshminarayanan_simple_2017}
Balaji Lakshminarayanan, Alexander Pritzel, and Charles Blundell. 2017.
\newblock \href {https://proceedings.neurips.cc/paper/2017/hash/9ef2ed4b7fd2c810847ffa5fa85bce38-Abstract.html} {Simple and scalable predictive uncertainty estimation using deep ensembles}.
\newblock In \emph{Advances in Neural Information Processing Systems 30: Annual Conference on Neural Information Processing Systems 2017, December 4-9, 2017, Long Beach, CA, {USA}}, pages 6402--6413.

\bibitem[{Lee et~al.(2018)Lee, Lee, Lee, and Shin}]{lee_simple_2018}
Kimin Lee, Kibok Lee, Honglak Lee, and Jinwoo Shin. 2018.
\newblock \href {https://proceedings.neurips.cc/paper/2018/hash/abdeb6f575ac5c6676b747bca8d09cc2-Abstract.html} {A simple unified framework for detecting out-of-distribution samples and adversarial attacks}.
\newblock In \emph{Advances in Neural Information Processing Systems 31: Annual Conference on Neural Information Processing Systems 2018, NeurIPS 2018, December 3-8, 2018, Montr{\'{e}}al, Canada}, pages 7167--7177.

\bibitem[{Lin et~al.(2022)Lin, Hilton, and Evans}]{lin2022teaching}
Stephanie Lin, Jacob Hilton, and Owain Evans. 2022.
\newblock \href {https://openreview.net/forum?id=8s8K2UZGTZ} {Teaching models to express their uncertainty in words}.
\newblock \emph{Transactions on Machine Learning Research}.

\bibitem[{Lyu et~al.(2023)Lyu, Havaldar, Stein, Zhang, Rao, Wong, Apidianaki, and Callison-Burch}]{fcot}
Qing Lyu, Shreya Havaldar, Adam Stein, Li~Zhang, Delip Rao, Eric Wong, Marianna Apidianaki, and Chris Callison-Burch. 2023.
\newblock \href {https://arxiv.org/abs/2301.13379} {Faithful chain-of-thought reasoning}.
\newblock \emph{ArXiv preprint}, abs/2301.13379.

\bibitem[{Madaan et~al.(2023)Madaan, Tandon, Gupta, Hallinan, Gao, Wiegreffe, Alon, Dziri, Prabhumoye, Yang, Welleck, Majumder, Gupta, Yazdanbakhsh, and Clark}]{self-refine}
Aman Madaan, Niket Tandon, Prakhar Gupta, Skyler Hallinan, Luyu Gao, Sarah Wiegreffe, Uri Alon, Nouha Dziri, Shrimai Prabhumoye, Yiming Yang, Sean Welleck, Bodhisattwa~Prasad Majumder, Shashank Gupta, Amir Yazdanbakhsh, and Peter Clark. 2023.
\newblock \href {https://arxiv.org/abs/2303.17651} {Self-refine: Iterative refinement with self-feedback}.
\newblock \emph{ArXiv preprint}, abs/2303.17651.

\bibitem[{Manakul et~al.(2023)Manakul, Liusie, and Gales}]{selfcheckgpt}
Potsawee Manakul, Adian Liusie, and Mark Gales. 2023.
\newblock \href {https://doi.org/10.18653/v1/2023.emnlp-main.557} {{S}elf{C}heck{GPT}: Zero-resource black-box hallucination detection for generative large language models}.
\newblock In \emph{Proceedings of the 2023 Conference on Empirical Methods in Natural Language Processing}, pages 9004--9017, Singapore. Association for Computational Linguistics.

\bibitem[{Miao et~al.(2020)Miao, Liang, and Su}]{miao_diverse_2020}
Shen-yun Miao, Chao-Chun Liang, and Keh-Yih Su. 2020.
\newblock \href {https://doi.org/10.18653/v1/2020.acl-main.92} {A diverse corpus for evaluating and developing {E}nglish math word problem solvers}.
\newblock In \emph{Proceedings of the 58th Annual Meeting of the Association for Computational Linguistics}, pages 975--984, Online. Association for Computational Linguistics.

\bibitem[{Mielke et~al.(2022)Mielke, Szlam, Dinan, and Boureau}]{mielke-etal-2022-reducing}
Sabrina~J. Mielke, Arthur Szlam, Emily Dinan, and Y-Lan Boureau. 2022.
\newblock \href {https://doi.org/10.1162/tacl_a_00494} {Reducing conversational agents{'} overconfidence through linguistic calibration}.
\newblock \emph{Transactions of the Association for Computational Linguistics}, 10:857--872.

\bibitem[{Nye et~al.(2021)Nye, Tessler, Tenenbaum, and Lake}]{nye2021improving}
Maxwell~I. Nye, Michael~Henry Tessler, Joshua~B. Tenenbaum, and Brenden~M. Lake. 2021.
\newblock \href {https://proceedings.neurips.cc/paper/2021/hash/d3e2e8f631bd9336ed25b8162aef8782-Abstract.html} {Improving coherence and consistency in neural sequence models with dual-system, neuro-symbolic reasoning}.
\newblock In \emph{Advances in Neural Information Processing Systems 34: Annual Conference on Neural Information Processing Systems 2021, NeurIPS 2021, December 6-14, 2021, virtual}, pages 25192--25204.

\bibitem[{OpenAI(2023)}]{openai2023gpt4}
OpenAI. 2023.
\newblock \href {http://arxiv.org/abs/2303.08774} {Gpt-4 technical report}.

\bibitem[{Papadopoulos et~al.(2001)Papadopoulos, Edwards, and Murray}]{papadopoulos2001confidence}
Georgios Papadopoulos, Peter~J Edwards, and Alan~F Murray. 2001.
\newblock Confidence estimation methods for neural networks: A practical comparison.
\newblock \emph{IEEE transactions on neural networks}, 12(6):1278--1287.

\bibitem[{Patel et~al.(2021)Patel, Bhattamishra, and Goyal}]{patel_are_2021}
Arkil Patel, Satwik Bhattamishra, and Navin Goyal. 2021.
\newblock \href {https://doi.org/10.18653/v1/2021.naacl-main.168} {Are {NLP} models really able to solve simple math word problems?}
\newblock In \emph{Proceedings of the 2021 Conference of the North American Chapter of the Association for Computational Linguistics: Human Language Technologies}, pages 2080--2094, Online. Association for Computational Linguistics.

\bibitem[{Portillo~Wightman et~al.(2023)Portillo~Wightman, Delucia, and Dredze}]{portillo-wightman-etal-2023-strength}
Gwenyth Portillo~Wightman, Alexandra Delucia, and Mark Dredze. 2023.
\newblock \href {https://doi.org/10.18653/v1/2023.trustnlp-1.28} {Strength in numbers: Estimating confidence of large language models by prompt agreement}.
\newblock In \emph{Proceedings of the 3rd Workshop on Trustworthy Natural Language Processing (TrustNLP 2023)}, pages 326--362, Toronto, Canada. Association for Computational Linguistics.

\bibitem[{Roy and Roth(2015)}]{roy_solving_2015}
Subhro Roy and Dan Roth. 2015.
\newblock \href {https://doi.org/10.18653/v1/D15-1202} {Solving general arithmetic word problems}.
\newblock In \emph{Proceedings of the 2015 Conference on Empirical Methods in Natural Language Processing}, pages 1743--1752, Lisbon, Portugal. Association for Computational Linguistics.

\bibitem[{Shridhar et~al.(2023)Shridhar, Jhamtani, Fang, Van~Durme, Eisner, and Xia}]{shridhar2023screws}
Kumar Shridhar, Harsh Jhamtani, Hao Fang, Benjamin Van~Durme, Jason Eisner, and Patrick Xia. 2023.
\newblock \href {https://arxiv.org/abs/2309.13075} {Screws: A modular framework for reasoning with revisions}.
\newblock \emph{ArXiv preprint}, abs/2309.13075.

\bibitem[{Shridhar et~al.(2022)Shridhar, Macina, El-Assady, Sinha, Kapur, and Sachan}]{shridhar2022automatic}
Kumar Shridhar, Jakub Macina, Mennatallah El-Assady, Tanmay Sinha, Manu Kapur, and Mrinmaya Sachan. 2022.
\newblock \href {https://aclanthology.org/2022.emnlp-main.277} {Automatic generation of socratic subquestions for teaching math word problems}.
\newblock In \emph{Proceedings of the 2022 Conference on Empirical Methods in Natural Language Processing}, pages 4136--4149, Abu Dhabi, United Arab Emirates. Association for Computational Linguistics.

\bibitem[{Sinha et~al.(2019)Sinha, Sodhani, Dong, Pineau, and Hamilton}]{sinha_clutrr_2019}
Koustuv Sinha, Shagun Sodhani, Jin Dong, Joelle Pineau, and William~L. Hamilton. 2019.
\newblock \href {https://doi.org/10.18653/v1/D19-1458} {{CLUTRR}: A diagnostic benchmark for inductive reasoning from text}.
\newblock In \emph{Proceedings of the 2019 Conference on Empirical Methods in Natural Language Processing and the 9th International Joint Conference on Natural Language Processing (EMNLP-IJCNLP)}, pages 4506--4515, Hong Kong, China. Association for Computational Linguistics.

\bibitem[{Tam et~al.(2022)Tam, Mascarenhas, Zhang, Kwan, Bansal, and Raffel}]{tam2022evaluating}
Derek Tam, Anisha Mascarenhas, Shiyue Zhang, Sarah Kwan, Mohit Bansal, and Colin Raffel. 2022.
\newblock \href {https://arxiv.org/abs/2211.08412} {Evaluating the factual consistency of large language models through summarization}.
\newblock \emph{ArXiv preprint}, abs/2211.08412.

\bibitem[{Touvron et~al.(2023)Touvron, Martin, Stone, Albert, Almahairi, Babaei, Bashlykov, Batra, Bhargava, Bhosale, Bikel, Blecher, Ferrer, Chen, Cucurull, Esiobu, Fernandes, Fu, Fu, Fuller, Gao, Goswami, Goyal, Hartshorn, Hosseini, Hou, Inan, Kardas, Kerkez, Khabsa, Kloumann, Korenev, Koura, Lachaux, Lavril, Lee, Liskovich, Lu, Mao, Martinet, Mihaylov, Mishra, Molybog, Nie, Poulton, Reizenstein, Rungta, Saladi, Schelten, Silva, Smith, Subramanian, Tan, Tang, Taylor, Williams, Kuan, Xu, Yan, Zarov, Zhang, Fan, Kambadur, Narang, Rodriguez, Stojnic, Edunov, and Scialom}]{LLaMA2}
Hugo Touvron, Louis Martin, Kevin Stone, Peter Albert, Amjad Almahairi, Yasmine Babaei, Nikolay Bashlykov, Soumya Batra, Prajjwal Bhargava, Shruti Bhosale, Daniel~M. Bikel, Lukas Blecher, Cristian~Cant{\'o}n Ferrer, Moya Chen, Guillem Cucurull, David Esiobu, Jude Fernandes, Jeremy Fu, Wenyin Fu, Brian Fuller, Cynthia Gao, Vedanuj Goswami, Naman Goyal, Anthony~S. Hartshorn, Saghar Hosseini, Rui Hou, Hakan Inan, Marcin Kardas, Viktor Kerkez, Madian Khabsa, Isabel~M. Kloumann, A.~V. Korenev, Punit~Singh Koura, Marie-Anne Lachaux, Thibaut Lavril, Jenya Lee, Diana Liskovich, Yinghai Lu, Yuning Mao, Xavier Martinet, Todor Mihaylov, Pushkar Mishra, Igor Molybog, Yixin Nie, Andrew Poulton, Jeremy Reizenstein, Rashi Rungta, Kalyan Saladi, Alan Schelten, Ruan Silva, Eric~Michael Smith, R.~Subramanian, Xia Tan, Binh Tang, Ross Taylor, Adina Williams, Jian~Xiang Kuan, Puxin Xu, Zhengxu Yan, Iliyan Zarov, Yuchen Zhang, Angela Fan, Melanie Kambadur, Sharan Narang, Aurelien Rodriguez, Robert Stojnic, Sergey Edunov, and
  Thomas Scialom. 2023.
\newblock \href {https://arxiv.org/abs/2307.09288} {Llama 2: Open foundation and fine-tuned chat models}.
\newblock \emph{ArXiv preprint}, abs/2307.09288.

\bibitem[{Wang et~al.(2023{\natexlab{a}})Wang, Liu, Yue, Tang, Zhang, Jiayang, Yao, Gao, Hu, Qi et~al.}]{wang2023survey}
Cunxiang Wang, Xiaoze Liu, Yuanhao Yue, Xiangru Tang, Tianhang Zhang, Cheng Jiayang, Yunzhi Yao, Wenyang Gao, Xuming Hu, Zehan Qi, et~al. 2023{\natexlab{a}}.
\newblock \href {https://arxiv.org/abs/2310.07521} {Survey on factuality in large language models: Knowledge, retrieval and domain-specificity}.
\newblock \emph{ArXiv preprint}, abs/2310.07521.

\bibitem[{Wang et~al.(2023{\natexlab{b}})Wang, Wei, Schuurmans, Le, Chi, Narang, Chowdhery, and Zhou}]{selfconsistency}
Xuezhi Wang, Jason Wei, Dale Schuurmans, Quoc~V Le, Ed~H. Chi, Sharan Narang, Aakanksha Chowdhery, and Denny Zhou. 2023{\natexlab{b}}.
\newblock \href {https://openreview.net/forum?id=1PL1NIMMrw} {Self-consistency improves chain of thought reasoning in language models}.
\newblock In \emph{The Eleventh International Conference on Learning Representations}.

\bibitem[{Wei et~al.(2022)Wei, Wang, Schuurmans, Bosma, brian ichter, Xia, Chi, Le, and Zhou}]{cot}
Jason Wei, Xuezhi Wang, Dale Schuurmans, Maarten Bosma, brian ichter, Fei Xia, Ed~H. Chi, Quoc~V Le, and Denny Zhou. 2022.
\newblock \href {https://openreview.net/forum?id=_VjQlMeSB_J} {Chain of thought prompting elicits reasoning in large language models}.
\newblock In \emph{Advances in Neural Information Processing Systems}.

\bibitem[{Xiong et~al.(2023)Xiong, Hu, Lu, Li, Fu, He, and Hooi}]{xiong2023can}
Miao Xiong, Zhiyuan Hu, Xinyang Lu, Yifei Li, Jie Fu, Junxian He, and Bryan Hooi. 2023.
\newblock \href {https://arxiv.org/abs/2306.13063} {Can llms express their uncertainty? an empirical evaluation of confidence elicitation in llms}.
\newblock \emph{ArXiv preprint}, abs/2306.13063.

\bibitem[{Yoo et~al.(2022)Yoo, Kim, Jang, and Kwak}]{yoo_detection_2022}
KiYoon Yoo, Jangho Kim, Jiho Jang, and Nojun Kwak. 2022.
\newblock \href {https://arxiv.org/abs/2203.01677} {Detection of {Word} {Adversarial} {Examples} in {Text} {Classification}: {Benchmark} and {Baseline} via {Robust} {Density} {Estimation}}.

\bibitem[{Zhou et~al.(2023)Zhou, Sch{\"a}rli, Hou, Wei, Scales, Wang, Schuurmans, Cui, Bousquet, Le, and Chi}]{zhou2023leasttomost}
Denny Zhou, Nathanael Sch{\"a}rli, Le~Hou, Jason Wei, Nathan Scales, Xuezhi Wang, Dale Schuurmans, Claire Cui, Olivier Bousquet, Quoc~V Le, and Ed~H. Chi. 2023.
\newblock \href {https://openreview.net/forum?id=WZH7099tgfM} {Least-to-most prompting enables complex reasoning in large language models}.
\newblock In \emph{The Eleventh International Conference on Learning Representations}.

\end{thebibliography}

\appendix


\section{Which Consistency Metrics Should I Use to Best Calibrate My Model?}
\label{decision-tree}
\begin{figure*}[h!]
\centering
\includegraphics[width=0.9\textwidth]{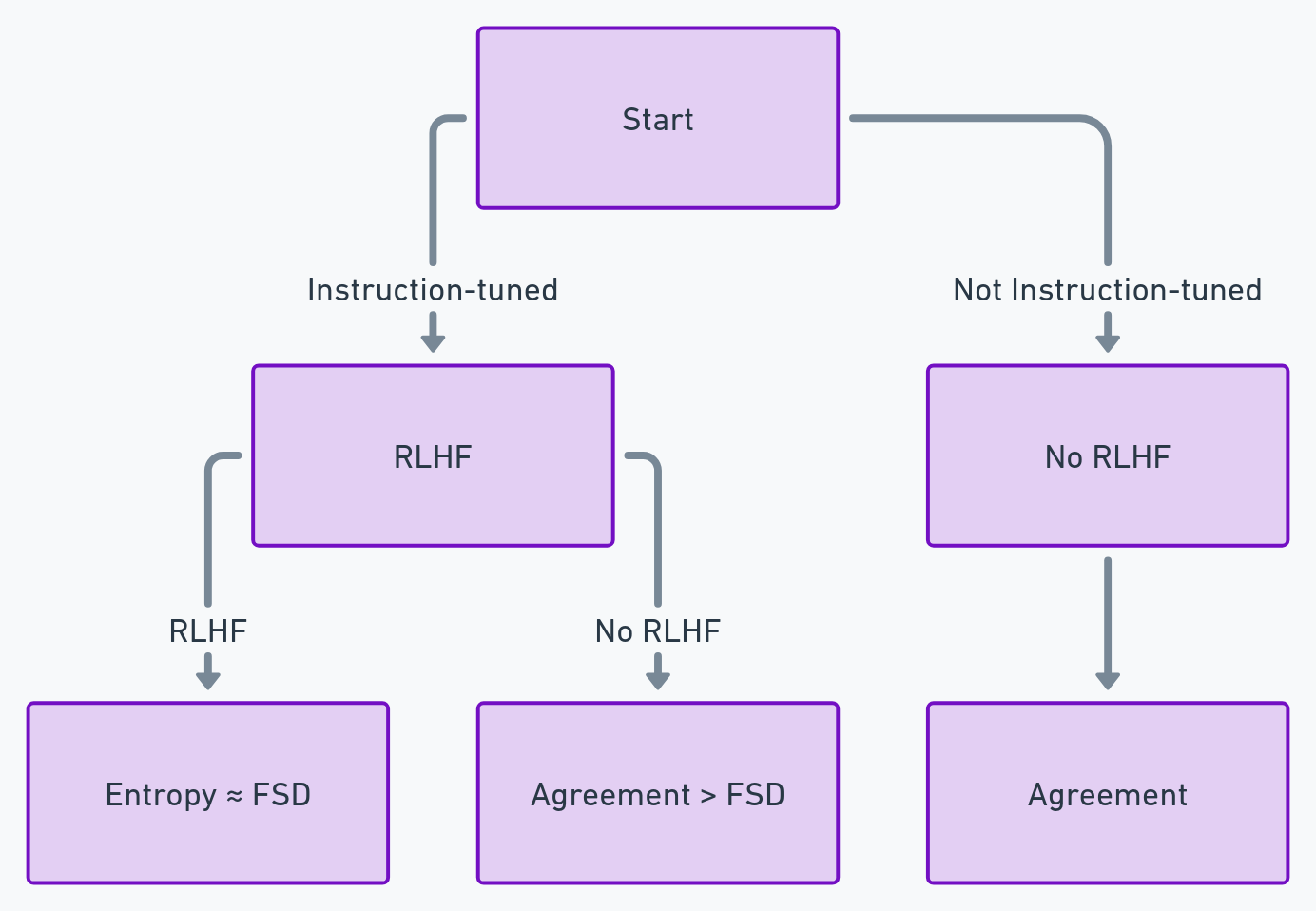}
\caption{A flow chart demonstrating the starting point of how to choose the consistency metric based on the model information in hand.}
\vspace{-0.13in}
\label{fig:prescription}
\end{figure*}

Depending on a model's specific characteristics, such as its exposure to instruction-tuning and RLHF, \autoref{fig:prescription} provides tailored recommendations for selecting appropriate consistency metrics for calibration. These suggestions are grounded in the insights derived from our analyses in Sections~\ref{sec:results} and ~\ref{sec:analysis}. For example, if the model has undergone both instruction-tuning and RLHF, an FSD-based or entropy-based consistency metric may be a good starting point. On the other hand, if the model has only been instruction-tuned without RLHF, an agreement-based consistency metric could be more suitable.

However, it is important to note that our research examined calibration properties in a somewhat limited scope, focusing on only four reasoning tasks across nine datasets. Additionally, certain comparisons (such as between instruction-tuned and non-instruction-tuned models) are based solely on a single pair of models (Mistral-7B vs. Mistral-7B-instruct). Consequently, our recommendations might not be universally applicable and should be applied judiciously."




\newpage

\section{Implementation Details}
\label{appendix:implementation_details}

\subsection{Closed-Source Models}
\label{appendix:implementation_details_close}

We use OpenAI Codex  (\texttt{code-davinci-002}, deprecated since Jan 4, 2024) \cite{chen_evaluating_2021}, GPT-3.5-turbo (\texttt{gpt-3.5-turbo-16k-0613}), and GPT-4 (\texttt{gpt-4-0613}) \cite{openai2023gpt4} through the Python API available at \url{platform.openai.com}, from Oct, 2023 to Feb, 2024. The inference cost per input query (with 40 samples of all five prompting strategies) is \$0 for all Codex models through the researcher access program, \$0.08 - \$0.13 for GPT-3.5-turbo, and \$0.61 - \$0.99 for GPT-4, depending on the dataset. The total cost of running inference on all 9 datasets is \$0 for Codex, around \$1,059 for GPT-3.5-turbo, and around \$7,942 for GPT-4.
The inference time on one input query (with 40 samples of all five prompting strategies) is 50 - 95 seconds with Codex under a rate limit of 150,000 tokens/minute, 39 - 74 seconds with GPT-3.5-turbo under 2,000,000 tokens/minute, and 83 - 157 seconds with GPT-4 under 300,000 tokens/minute, also depending on the dataset. The total time for running inference on all 9 datasets is 8.3 days for Codex, 6.4 days for GPT-3.5-turbo, and 13.8 days for GPT-4.

We use the following hyper-parameters throughout all experiments:
\begin{itemize}
    \item \textbf{temperature}: 0.0 for greedy decoding, 0.4 for self-consistent decoding;
    \item \textbf{max\_tokens}: 1000;
    \item  \textbf{n}: 1 for greedy decoding, 40 for self-consistent decoding;
    \item \textbf{frequency\_penalty}: 0;
    \item \textbf{presence\_penalty}: 0.
\end{itemize}

Any unspecified hyper-parameters are set to the default value on \url{https://platform.openai.com/docs/api-reference/completions/create} and \url{https://platform.openai.com/docs/api-reference/chat}.

\subsection{Open-Source Models}
\label{appendix:implementation_details_open}

We use LLaMA (7B/13B/70B) \cite{LLaMA2} and Mistral (7B/7B-instruct) \cite{jiang2023mistral} as the open-source models in our experiments. We used Nvidia A100 80GB GPUs to generate output for all open-source models. The LLaMA-70B model used 2 GPUs for each inference, while all other models used a single A100 GPU. The checkpoints and tokenizers were loaded from their respective official repositories on HuggingFace (\texttt{meta-llama} for LLaMA models and \texttt{mistralai} for Mistral models). The hyperparameters were kept the same as in the closed-source models for a fair comparison. On average, each inference took less than a second for the standard strategy, 3-4 seconds for CoT and LtM, and 5-6 seconds for FCoT and PoT for all 7B models (LLaMA-7B, Mistral-7B, and Mistral-7B-instruct). The LLaMA-13B took 1.5 times longer on average and the LLaMA-70B took 4 times longer on average.  In terms of GPU hours (Nvidia A100 80GB), the LLaMA-7B, Mistral-7B, and Mistral-7B-instruct models took about 9 hours for LtM and CoT strategies, 4.5 hours for Standard, and 13 hours for PoT and FCoT strategies. In total, it took approximately 50 hours for each of the LLaMA-7B, Mistral-7B, and Mistral-7B-instruct models to run experiments for all strategies across all datasets. For LLaMA-13B it took about 75 hours and for LLaMA-70B about 200 hours. Due to the formidable computation cost of up to 425 hours, we have not finished running all baselines for open-source models yet.

\subsection{Case Study Details}
 In the discrimination experiment, we tune an optimal threshold $\theta$ for each calibration method on a development set with 100 samples. The range of $\theta$ is from 0.0 to 0.9 with a step size of 0.05 and from 0.9 to 1.0 with a step size of 0.01. We find the best threshold with the highest discrimination Macro-F1 score on the development set, and use this threshold on the test set.

\section{Baseline Details}
\label{appendix:baselines} 

We describe the details on how we implement the baselines in Section~\ref{sec:implemetation_details}. Given an input $x$ and the most-voted answer $\Bar{a}$, we want to get an estimated confidence score $\texttt{conf}(x, \Bar{a})$ of the answer being correct from each calibration method.

\paragraph{Raw logits (logit).} We measure the confidence as the exponential of the average log probability of all tokens in a sample reasoning chain $\hat{s}_{\Bar{a}}$ that results in the answer $\Bar{a}$. This is equivalent to the reciprocal of the perplexity of the reasoning chain, or $\frac{1}{\text{PPL}(\hat{s}_{\Bar{a}})}$.

\paragraph{P(True).} We prompt the model to examine the correctness of its generated answer $\Bar{a}$ and reasoning chain $\hat{s}_{\Bar{a}}$ with the following prompt:

{\footnotesize
\begin{lstlisting}[frame=none]
Q: {QUERY}
A: {REASONING_CHAIN}
Answer: {ANSWER}
Is the above answer correct? (Yes/No):
\end{lstlisting}
}

We then take the normalized probability of the token ``Yes'' as the confidence, or $\frac{P(\text{Yes})}{P(\text{Yes})+P(\text{No})}$, where $P()$ is the probability assigned to a token by the LM, considering both its uppercase and lowercase variants.

We do not use the original prompt from \citet{kadavath2022language} that uses ``True/False'' instead of ``Yes/No'', because we find that the model sometimes have difficulty outputting the token in the required format in a 0-shot setting. 

We implement P(True) under both 0-shot and 8-shot prompting in our experiments. In the 0-shot setting (\textbf{ptrue\textsubscript{0-shot}}), we directly prompt the model with the above prompt. In the 8-shot setting (\textbf{ptrue\textsubscript{8-shot}}), we additionally show 8 exemplars in the same format randomly sampled from the development set, with 4 correct (``Yes'') and 4 incorrect (``No'') predictions in random order. The full prompts can be found in the Supplementary Materials.

\paragraph{Verbalized Confidence.} Similar to P(True), we prompt the model to examine the query and its generated answer $\Bar{a}$ and reasoning chain $\hat{s}_{\Bar{a}}$. However, we now ask it to directly verbalize its confidence either as a percentage (\textbf{verb\textsubscript{percent}}):

{\footnotesize
\begin{lstlisting}[frame=none]
Q: {QUERY}
A: {REASONING_CHAIN}
Answer: {ANSWER}
How confident are you in the above answer
(0-100%)?:
\end{lstlisting}
}

or as a linguistic expression (\textbf{verb\textsubscript{ling}}):

{\footnotesize
\begin{lstlisting}[frame=none]
Q: {QUERY}
A: {REASONING_CHAIN}
Answer: {ANSWER}
How confident are you in the above answer?
(choose from "Almost no chance", "Highly 
unlikely", "Unlikely", "Probably not", 
"About even", "Better than even", 
"Likely", "Probably", "Highly likely", 
"Almost certain"):
\end{lstlisting}
}

where the linguistic expressions above are deterministically mapped to a percentage from 5\% to 95\% with a step size of 5\%, in the listed order.

Finally, we take the predicted percentage or the linguistic expression mapped to a percentage as the verbalized confidence level. For both verb\textsubscript{percent} and verb\textsubscript{ling}, we use 0-shot prompting, since it is technically impossible to know the true ``confidence'' of a single prediction. Also, our experimental setup assumes a post-hoc setting, where no additional data is available for tuning a mapping from linguistic expressions to percentages.

\section{Dataset Details}
\label{appendix:dataset_details}

\begin{table*}[h]
\centering
\scalebox{0.78}{
\begin{tabular}{p{1.5cm}p{1.8CM}>{\raggedleft\arraybackslash}p{1.1cm}>{\raggedleft\arraybackslash}p{1cm}p{12.5cm}}
    \toprule \textbf{Domain} & \textbf{Dataset} & \bf{\# Shot}  & \bf{\# Test} & \textbf{Example}  \\
    \midrule \multirow{13}{1.5cm}{Math Word Problems} & GSM8K & 8 & 1,319 & Q: Natalia sold clips to 48 of her friends in April, and then she sold half as many clips in May. How many clips did Natalia sell altogether in April and May? \newline A: \texttt{72} \\
    & SVAMP & 8 & 1,000 & Q: Each pack of dvds costs 76 dollars. If there is a discount of 25 dollars on each pack. How much do you have to pay to buy each pack? \newline A: \texttt{51} \\
    & MultiArith & 8 & 600 &  Q: For Halloween Debby and her sister combined the candy they received. Debby had 32 pieces of candy while her sister had 42. If they ate 35 pieces the first night, how many pieces do they have left? \newline A: \texttt{39} \\
    & ASDiv & 8 & 2,096 &  Q: Seven red apples and two green apples are in the basket. How many apples are in the basket? \newline A: \texttt{9} \\
    \midrule
    \multirow{6}{1.5cm}{Multi-hop \newline QA} & StrategyQA & 6 & 2,290 & Q: Did Aristotle use a laptop? \newline A: \texttt{False} \\
    & Date \newline Understanding & 10 & 359 & Q: Yesterday was April 30, 2021. What is the date tomorrow in MM/DD/YYYY? \newline A: \texttt{``05/02/2021''} \\
    & Sports \newline Understanding & 10 & 977 &  Q: Is the following sentence plausible: ``Lebron James hit the turnaround jumper''? \newline A: \texttt{True} \\
    \midrule
    \multirow{3}{1.5cm}{Planning} & SayCan & 7 & 103 & Q: Could you get me a drink with caffeine? \newline A: \texttt{``1.find(redbull)\ 2.pick(redbull)\ 3.find(user)\ 4.put(redbull)\ 5.done().''} \\
    \midrule
    \multirow{4}{1.5cm}{Relational \newline Inference} & CLUTRR & 8 & 1,042 & Q: [Carlos] is [Clarence]'s brother. [Carlos] and his sister, [Annie], went shopping. [Annie] asked her mom [Valerie] if she wanted anything, but [Valerie] said no. How is [Valerie] related to [Clarence]? \newline A: \texttt{``mother''} \\
    \bottomrule    
\end{tabular}
}
\caption{Datasets used for evaluation. ``\# Shot'' stands for the number of few-shot examples in the prompt (following \citet{cot}) and ``\# Test'' stands for the number of test examples.}
\label{table:datasets}
 \vspace{-0.1in}
\end{table*}

\subsection{Dataset Description}

\paragraph{Math Word Problems (MWP).} Given a math problem written in NL, the goal is to derive the answer as a real-valued number. We follow \citet{cot} and consider the following MWP benchmarks: \textbf{GSM8K} \cite{cobbe_training_2021}, \textbf{SVAMP} \cite{patel_are_2021}, \textbf{MultiArith} \cite{roy_solving_2015}, and \textbf{ASDiv} \cite{miao_diverse_2020}. We use the same prompt for all these datasets.

\paragraph{Multi-hop QA.} Given a complex question $Q$ that involves multiple steps of reasoning, we want to obtain the answer as a Boolean value or string value variable. We consider three datasets: \textbf{StrategyQA} \cite{geva_did_2021}, a dataset of science questions that require an implicit multi-step strategy to answer; \textbf{Date Understanding} from BIG-bench \cite{big-bench_collaboration_beyond_2021}, which involves questions about inferring a date by performing computation on relative periods of time; and \textbf{Sports Understanding} from BIG-bench, which involves deciding whether an artificially constructed statement related to sports is plausible or not.

\paragraph{Planning.} We use the \textbf{SayCan} dataset \cite{ahn_as_2022}, which assumes a scenario of a robot operating in a kitchen, helping the user with household tasks, e.g., ``I spilled my coke on the table; can you throw it away and bring me something to clean up?''. There are a number of locations and objects that the robot can interact with. The robot can only perform a fixed set of actions, including \texttt{find}, \texttt{pick}, and \texttt{put}. The task is to map a user query in NL to a plan of predefined actions performed on the objects and/or locations.

\paragraph{Relational inference.} We use the \textbf{CLUTRR} dataset. Given a short story about family relationships among multiple people, the goal is to infer the relationship between two specific people. The dataset has multiple splits based on the number of intermediate steps $K$ required to reach the answer. We construct the prompt using 8 exemplars with $K \in \{2,3\}$, and test the models on the remaining examples with $K$ up to 10.

\subsection{Statistics}

We show the dataset details in Table~\ref{table:datasets}, including the statistics, the number of few-shot exemplars used in the prompt, and example inputs and outputs.

\subsection{URLs and Licenses}

We use the same distribution of datasets following \citet{cot}:

\paragraph{Math Word Problems}
\begin{itemize}
    \item GSM8K \citep{cobbe_training_2021}: \url{https://github.com/openai/grade-school-math}, MIT license: \url{https://github.com/openai/grade-school-math/blob/master/LICENSE}.
    \item SVAMP \citep{patel_are_2021}: \url{https://github.com/arkilpatel/SVAMP}, MIT license: \url{https://github.com/arkilpatel/SVAMP/blob/main/LICENSE}.
    \item MultiArith \citep{roy_solving_2015}, license: CC BY 4.0.
    \item ASDiv \citep{miao_diverse_2020}: \url{https://github.com/chaochun/nlu-asdiv-dataset}.
    
\end{itemize}

\paragraph{Multi-hop QA}
\begin{itemize}
    \item StrategyQA \citep{geva_did_2021}: we use the open-domain setting (question-only set) from \cite{big-bench_collaboration_beyond_2021}: \url{https://github.com/google/BIG-bench/tree/main/bigbench/benchmark_tasks/strategyqa}.
    \item Date Understanding and Sports Understanding from BIG-Bench \citep{big-bench_collaboration_beyond_2021}: Apache License v.2: \url{https://github.com/google/BIG-bench/blob/main/LICENSE}.
\end{itemize}

\paragraph{Planning}
\begin{itemize}
    \item SayCan \citep{ahn_as_2022}: SayCan dataset can be accessed at \url{https://say-can.github.io/} under CC BY 4.0 license.
\end{itemize}

\paragraph{Relational Reasoning}
\begin{itemize}
    \item CLUTRR \citep{sinha_clutrr_2019}: \url{https://github.com/facebookresearch/clutrr}, license: \url{https://github.com/facebookresearch/clutrr/blob/main/LICENSE}. We obtain the publicly distributed version available at \url{https://drive.google.com/file/d/1SEq_e1IVCDDzsBIBhoUQ5pOVH5kxRoZF/view}, specifically the \texttt{data\_089907f8} split.
\end{itemize}

We use all the above datasets for research purposes only, consistent with their intended use. We use the same preprocessed version and train/dev/test split of the datasets as \citet{fcot}.

\section{Additional Results}
\label{appendix:additional_results}

\subsection{End Task Accuracy}

Table~\ref{table:accuracy} shows the accuracy of each LM and prompting strategy, averaged over all datasets.

\begin{table}[h]
    \centering
    \scalebox{0.75}{
    \begin{tabular}{l c c c c c}
    \toprule

        \textbf{Model} & \textbf{Standard} & \textbf{CoT} & \textbf{LtM} & \textbf{PoT}  & \textbf{FCoT}\\ 
        \midrule
         Codex & 57.1 &	81.3 & 74.3 &	80.0 &	\textbf{83.4} \\ 
         GPT-3.5.turbo & 64.9&	\textbf{77.6} &	\textbf{77.6} &	72.5 &	76.8\\ 
         GPT-4 & 79.3	&88.3	&87.3	&84.4	&\textbf{90.9}\\ 
         LLaMA-7B & 40.1	&\textbf{56.4}	&46.0	&47.4	&50.2\\
         LLaMA-13B&43.2	&\textbf{66.8}	&58.2	&56.9	&62.2\\
         LLaMA-70B& 58.0	&\textbf{82.0}	&73.3	&73.6	&77.0\\
         Mistral-7B& 49.9	&\textbf{73.5}	&61.9	&66.5	&71.2\\
         Mistral-7B-instruct & 43.7	&63.6	&56.0	&60.0	&\textbf{67.1}\\
        \bottomrule
    \end{tabular}
    }
    \vspace{-0.03in}
    \caption{Accuracy (in \%) averaged across all datasets for various LLMs. The best accuracy for a given model is highlighted in \textbf{bold}.}
    \label{table:accuracy}
\end{table}

\subsection{Comparing Consistency Merics}

Table~\ref{table:3_consistency_significance} compares the efficacy of three consistency metrics in terms of Brier Score averaged over all datasets and prompting strategies, with significance level. We can observe that Codex and all open-source models prefer agreement as the best or second-best (not significantly different from the best) consistency measure. GPT-3.5-turbo and GPT-4 prefer entropy and FSD, which have the same performance considering statistical significance ($p \geq 0.05$).

\begin{table}[!t]
    \centering
    \addtolength{\tabcolsep}{-3pt}
    \scalebox{0.8}{
    \begin{tabular}{p{3cm}|>{\raggedleft\arraybackslash}p{1.8cm}>{\raggedleft\arraybackslash}p{1.8cm}>{\raggedleft\arraybackslash}p{1.8cm}}
    \toprule
        \textbf{LM} & \multicolumn{3}{c}{\textbf{Consistency Metrics}}\\
        & \textbf{entropy} & \textbf{agree} & \textbf{FSD}  \\ 
        \midrule
        Codex & .175† & \textbf{.151}\phantom{†} & .159†  \\ 
        GPT-3.5-turbo & \textbf{.205}\phantom{†} & .221† & .207\phantom{†} \\ 
        GPT-4 & .116\phantom{†} & .119† & \textbf{.114}\phantom{†} \\ 
        LLaMA-7B & .241† & \textbf{.232}\phantom{†} & .235† \\ 
        LLaMA-13B & .222† & \textbf{.204}\phantom{†} & .211† \\ 
        LLaMA-70B & .182† & \textbf{.154}\phantom{†} & .165† \\ 
        Mistral-7B & .205† & \textbf{.183}\phantom{†} & .191† \\ 
        Mistral-7B-instruct & .220† & .216\phantom{†} & \textbf{.215}\phantom{†} \\ 
        \bottomrule
    \end{tabular}
    }
    \caption{Overall Brier Score ($\downarrow$) of three consistency metrics averaged across all datasets and prompting strategies. † indicates that the current metric is significantly worse ($p<0.05$) than the best-performing metric (in bold).}
    \label{table:3_consistency_significance}
\end{table}

\subsection{Calibration Results on All Datasets}
\label{appendix:calib_results_all_datasets}

Table~\ref{table:results_5.1_close_all_datasets} and Table~\ref{table:results_5.1_open_all_datasets} compare the Brier Score of all calibration methods for closed-source and open-source models on all 9 datasets. 

\begin{figure*}[t!]
\centering
\includegraphics[width=0.9\textwidth]{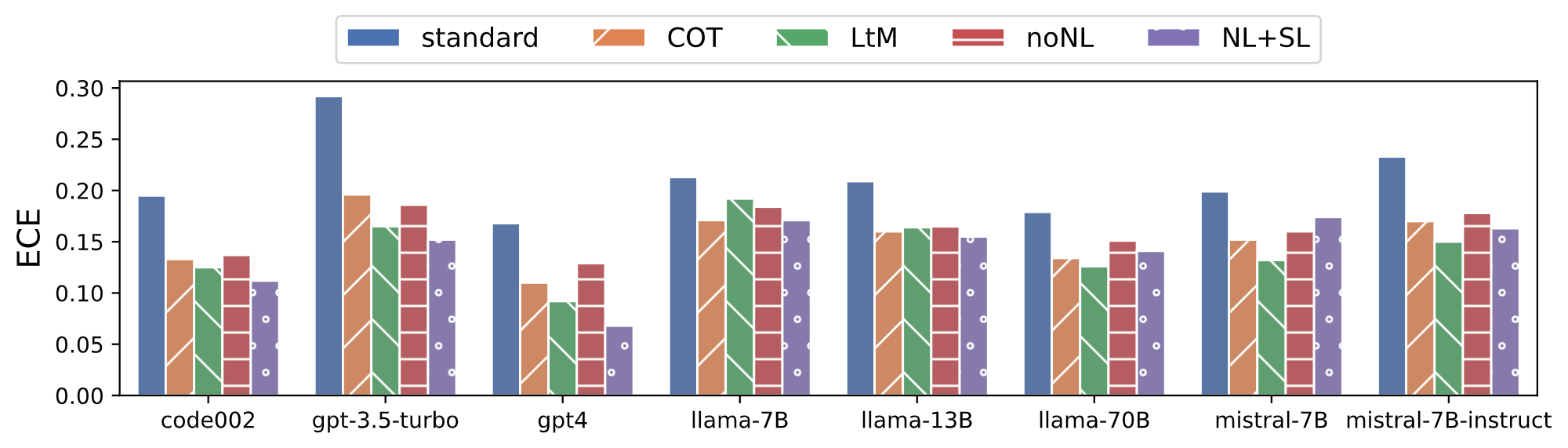}
\caption{ECE score ($\downarrow$) for each prompting strategy, averaged across all datasets and consistency metrics.}
\label{fig:results_compare_prompts_ece}
\end{figure*}

Table~\ref{table:results_ece} and Figure~\ref{fig:results_compare_prompts_ece} show the ECE of all calibration methods for all models on all domains. We can observe that they exhibit similar trends as the Brier Score.

\begin{table*}[!b]
    \centering
    \scalebox{0.75}{
    \begin{tabular}{p{2.5cm}|>{\raggedleft\arraybackslash}p{1.8cm}>{\raggedleft\arraybackslash}p{1.8cm}>{\raggedleft\arraybackslash}p{1.8cm}|>{\raggedleft\arraybackslash}p{1.8cm}>{\raggedleft\arraybackslash}p{1.8cm}>{\raggedleft\arraybackslash}p{1.8cm}>{\raggedleft\arraybackslash}p{1.8cm}>{\raggedleft\arraybackslash}p{1.8cm}}
    \toprule
        \textbf{Dataset} & \multicolumn{3}{c|}{\textbf{Consistency Metrics}} & \multicolumn{5}{c}{\textbf{Baselines}} \\

         & \textbf{entropy} & \textbf{agreement} & \textbf{FSD} & \textbf{verb\textsubscript{ling}} & \textbf{verb\textsubscript{percent}} & \textbf{logit} & \textbf{ptrue\textsubscript{0-shot}} & \textbf{ptrue\textsubscript{8-shot}} \\ \midrule
        & \multicolumn{8}{c}{LM: \texttt{Codex}} \vspace{0.05cm} \\\hline

        ASDiv & .099 & \textbf{.090} & .095 & .205 & .190 & .150 & .159 & .120 \\ 
        GSM8K & .189 & \textbf{.158} & .177 & .252 & .377 & .262 & .248 & .188 \\ 
        Multi & .103 & \textbf{.085} & .089 & .187 & .162 & .135 & .106 & .117 \\ 
        SVAMP & .126 & \textbf{.103} & .114 & .193 & .173 & .139 & .145 & .148 \\ 
        Sport & .071 & .068 & \textbf{.062} & .346 & .075 & .067 & .103 & .200 \\ 
        Date & .174 & \textbf{.159} & .162 & .210 & .251 & .219 & .191 & .197 \\ 
        StrategyQA & .353 & \textbf{.213} & .256 & .305 & .316 & .257 & .271 & .220 \\ 
        CLUTRR & \textbf{.288} & .369 & .327 & .296 & .536 & .506 & .330 & .232 \\ 
        SayCan & .175 & \textbf{.115} & .152 & .243 & .159 & .145 & .135 & .190 \\ 
        average & .175 & \textbf{.151} & .159 & .249 & .249 & .209 & .188 & .179 \\ 
        \midrule
        
        & \multicolumn{8}{c}{LM: \texttt{GPT-3.5-turbo}} \vspace{0.05cm} \\\hline
        
        ASDiv & \textbf{.194} & .224 & \textbf{.194} & .223 & .213 & n/a & n/a & n/a \\ 
        GSM8K & .184 & .196 & \textbf{.183} & .260 & .338 & n/a & n/a & n/a \\ 
        MultiArith & .044 & .041 & \textbf{.039} & .101 & .065 & n/a & n/a & n/a \\ 
        SVAMP & \textbf{.108} & .115 & \textbf{.108} & .164 & .155 & n/a & n/a & n/a \\ 
        Sport & \textbf{.089} & .101 & .095 & .326 & .151 & n/a & n/a & n/a \\ 
        Date & \textbf{.266} & .292 & .280 & .316 & .348 & n/a & n/a & n/a \\ 
        StrategyQA & .393 & .411 & .376 & \textbf{.329} & .342 & n/a & n/a & n/a \\ 
        CLUTRR & \textbf{.429} & .482 & .465 & .450 & .509 & n/a & n/a & n/a \\ 
        SayCan & .137 & \textbf{.126} & \textbf{.126} & .267 & .341 & n/a & n/a & n/a \\ 
        average & \textbf{.205} & .221 & .207 & .271 & .273 & n/a & n/a & n/a \\ 
        \midrule
        
        & \multicolumn{8}{c}{LM: \texttt{GPT-4}} \vspace{0.05cm} \\\hline
        
        ASDiv & \textbf{.090} & .103 & \textbf{.090} & .091 & .095 & n/a & n/a & n/a \\ 
        GSM8K & \textbf{.083} & .099 & .087 & .132 & .144 & n/a & n/a & n/a \\ 
        MultiArith & .013 & \textbf{.010} & .011 & .015 & .013 & n/a & n/a & n/a \\ 
        SVAMP & \textbf{.047} & .050 & \textbf{.047} & .058 & .063 & n/a & n/a & n/a \\ 
        Sport & .033 & \textbf{.031} & \textbf{.031} & .160 & .100 & n/a & n/a & n/a \\ 
        Date & .063 & .069 & \textbf{.061} & .073 & .076 & n/a & n/a & n/a \\ 
        StrategyQA & .230 & .205 & .207 & \textbf{.195} & .220 & n/a & n/a & n/a \\ 
        CLUTRR & .392 & .443 & .416 & \textbf{.386} & .435 & n/a & n/a & n/a \\ 
        SayCan & .092 & \textbf{.065} & .079 & .279 & .481 & n/a & n/a & n/a \\ 
        average & .116 & .119 & \textbf{.114} & .154 & .181 & n/a & n/a & n/a \\ 
        \bottomrule
    \end{tabular}
    }
    \caption{Brier Score ($\downarrow$) for closed-source LMs on all datasets, averaged across five prompting strategies. The best scores are \textbf{in bold}.}
    \label{table:results_5.1_close_all_datasets}
\end{table*}

\begin{table*}[!ht]
    \centering
    \scalebox{0.8}{
    \begin{tabular}{p{2.5cm}|>{\raggedleft\arraybackslash}p{1.8cm}>{\raggedleft\arraybackslash}p{1.8cm}>{\raggedleft\arraybackslash}p{1.8cm}|>{\raggedleft\arraybackslash}p{1.8cm}}
    \toprule
        \textbf{Domain} & \multicolumn{3}{c|}{\textbf{Consistency Metrics}} & \multicolumn{1}{c}{\textbf{Baselines}} \\

        & \textbf{entropy} & \textbf{agree} & \textbf{FSD} & \textbf{logit} \\ 
        \midrule
         & \multicolumn{4}{c}{LM: \texttt{LLaMA-7B}} \vspace{0.05cm} \\\hline
         
        ASDiv & .164 & \textbf{.155} & .158 & .411 \\ 
        GSM8K & \textbf{.137} & .166 & .148 & .713 \\ 
        MultiArith & .232 & \textbf{.231} & .241 & .544 \\ 
        SVAMP & .211 & \textbf{.195} & .211 & .455 \\ 
        Sport & .260 & \textbf{.221} & .232 & .272 \\ 
        Date & \textbf{.216} & .267 & .235 & .526 \\ 
        StrategyQA & .390 & \textbf{.265} & .301 & .408 \\ 
        CLUTRR & \textbf{.290} & .370 & .323 & .633 \\ 
        SayCan & .267 & \textbf{.214} & .269 & .307 \\ 
        average & .241 & \textbf{.232} & .235 & .474 \\ 
        \midrule
        
         & \multicolumn{4}{c}{LM: \texttt{LLaMA-13B}} \vspace{0.05cm} \\\hline
         
        ASDiv & .144 & \textbf{.135} & .136 & .334 \\ 
        GSM8K & \textbf{.177} & .179 & .181 & .591 \\ 
        MultiArith & .232 & \textbf{.198} & .222 & .395 \\ 
        SVAMP & .205 & \textbf{.180} & .200 & .365 \\ 
        Sport & .170 & .153 & \textbf{.151} & .194 \\ 
        Date & \textbf{.181} & .206 & .185 & .402 \\ 
        StrategyQA & .383 & \textbf{.241} & .285 & .371 \\ 
        CLUTRR & \textbf{.298} & .353 & .320 & .596 \\ 
        SayCan & .209 & \textbf{.190} & .220 & .250 \\ 
        average & .222 & \textbf{.204} & .211 & .389 \\ 
        \midrule
        
        & \multicolumn{4}{c}{LM: \texttt{LLaMA-70B}} \vspace{0.05cm} \\\hline
         
        ASDiv & .107 & \textbf{.099} & .101 & .209 \\ 
        GSM8K & .201 & \textbf{.168} & .187 & .375 \\ 
        MultiArith & .134 & \textbf{.108} & .113 & .188 \\ 
        SVAMP & .145 & \textbf{.112} & .129 & .183 \\ 
        Sport & .053 & \textbf{.041} & .044 & .046 \\ 
        Date & .167 & \textbf{.166} & .168 & .262 \\ 
        StrategyQA & .338 & \textbf{.192} & .239 & .289 \\ 
        CLUTRR & \textbf{.287} & .347 & .309 & .534 \\ 
        SayCan & .206 & \textbf{.156} & .191 & .179 \\ 
        average & .182 & \textbf{.154} & .165 & .252 \\ 
        \midrule
        
         & \multicolumn{4}{c}{LM: \texttt{Mistral-7B}} \vspace{0.05cm} \\\hline
         
        ASDiv & \textbf{.122} & \textbf{.112} & .113 & .269 \\ 
        GSM8K & .197 & \textbf{.188} & .196 & .491 \\ 
        MultiArith & .189 & \textbf{.160} & .171 & .320 \\ 
        SVAMP & .176 & \textbf{.141} & .162 & .250 \\ 
        Sport & .116 & \textbf{.085} & .090 & .109 \\ 
        Date & \textbf{.178} & .202 & .187 & .373 \\ 
        StrategyQA & .363 & \textbf{.239} & .276 & .343 \\ 
        CLUTRR & \textbf{.283} & .334 & .307 & .555 \\ 
        SayCan & .225 & \textbf{.186} & .217 & .207 \\ 
        average & .205 & \textbf{.183} & .191 & .324 \\ 
        \midrule
        
        & \multicolumn{4}{c}{LM: \texttt{Mistral-7B-instruct}} \vspace{0.05cm} \\\hline
         
        ASDiv & .130 & .127 & \textbf{.124} & .288 \\ 
        GSM8K & .193 & \textbf{.191} & .194 & .508 \\ 
        MultiArith & .191 & \textbf{.164} & .180 & .319 \\ 
        SVAMP & .166 & \textbf{.147} & .158 & .274 \\ 
        Sport & .207 & .196 & \textbf{.194} & .222 \\ 
        Date & \textbf{.220} & .244 & .227 & .497 \\ 
        StrategyQA & .334 & \textbf{.269} & .284 & .366 \\ 
        CLUTRR & \textbf{.306} & .397 & .347 & .648 \\ 
        SayCan & .228 & \textbf{.209} & .229 & .332 \\ 
        average & .220 & .216 & \textbf{.215} & .384 \\ 
        \bottomrule
    \end{tabular}
    }
    \vspace{-0.03in}
    \caption{Brier Score ($\downarrow$) for closed-source LMs on all datasets, averaged across five prompting strategies. The best scores are \textbf{in bold}.}
    \label{table:results_5.1_open_all_datasets}
\end{table*}

\begin{table*}[!ht]
    \centering
    \scalebox{0.75}{
    \begin{tabular}{p{1.5cm}|>{\raggedleft\arraybackslash}p{1.8cm}>{\raggedleft\arraybackslash}p{1.8cm}>{\raggedleft\arraybackslash}p{1.8cm}|>{\raggedleft\arraybackslash}p{1.8cm}>{\raggedleft\arraybackslash}p{1.8cm}>{\raggedleft\arraybackslash}p{1.8cm}>{\raggedleft\arraybackslash}p{1.8cm}>{\raggedleft\arraybackslash}p{1.8cm}}
    \toprule
        \textbf{Domain} & \multicolumn{3}{c|}{\textbf{Consistency Metrics}} & \multicolumn{5}{c}{\textbf{Baselines}} \\

         & \textbf{entropy} & \textbf{agreement} & \textbf{FSD} & \textbf{verb\textsubscript{ling}} & \textbf{verb\textsubscript{percent}} & \textbf{logit} & \textbf{ptrue\textsubscript{0-shot}} & \textbf{ptrue\textsubscript{8-shot}} \\ \midrule
        & \multicolumn{8}{c}{LM: \texttt{Codex}} \vspace{0.05cm} \\
        \hline
         MWP & .132	& \textbf{.077} & .104 &	.237 &	.225 & .156 & .142	& .108 \\ 
        MHQA & .188 & \textbf{.090}	& .119	&.272	& .214	&.152 & n/a & .167 \\ 
        Plan. & .203	&\textbf{.101}	&.159	&.322	&.159	&.106	&.117	&.248 \\ 
        Relation. & .228 &	.368	&.294	&.214	&.536	&.512	&.313	&\textbf{.175} \\ 
        average & .169	&\textbf{.117}	&.136	&.256	&.249	&.189	&.166	&.151 \\ 
        \midrule
        
        & \multicolumn{8}{c}{LM: \texttt{GPT-3.5-turbo}} \vspace{0.05cm} \\
        \hline
        MWP & \textbf{.118}	&.121	&.115	&.208	&.193 & n/a & n/a & n/a \\ 
        MHQA & \textbf{.230}	&.246	&.233	&.321	&.277 & n/a & n/a & n/a \\ 
        Plan. & .154	&\textbf{.119}	&.128	&.351	&.357 & n/a & n/a & n/a \\ 
        Relation. & \textbf{.426}	&.505	&.471	&.449	&.519 & n/a & n/a & n/a \\ 
        average & \textbf{.193}	&.205	&.195	&.289	&.275 & n/a & n/a & n/a \\ 
        \midrule
        
        & \multicolumn{8}{c}{LM: \texttt{GPT-4}} \vspace{0.05cm} \\
        \hline
        MWP & .056	&.064	&.055	&\textbf{.053}	&.079 & n/a & n/a & n/a \\ 
        MHQA & .104	&\textbf{.089}	&.090	&.139	&.126 & n/a & n/a & n/a \\ 
        Plan. & .109	&\textbf{.061}	&.084	&.351	&.484 & n/a & n/a & n/a \\ 
        Relation. & .387	&.454	&.415	&\textbf{.381}	&.435 & n/a & n/a & n/a \\ 
        average & .114	&.115	&\textbf{.110}	&.151	&.179 & n/a & n/a & n/a \\ 
        \midrule
       
        & \multicolumn{8}{c}{LM: \texttt{LLaMA-7B}} \vspace{0.05cm} \\
        \hline
        MWP & .138	&\textbf{.130}	&.141	& - & - & .548 & - & - \\ 
        MHQA & .237	&\textbf{.189}	&.197	& - & - & .400 & - & - \\ 
        Plan. & .256	&\textbf{.164}	&.259	& - & - & .302 & - & - \\ 
        Relation.& \textbf{.214}	&.359	&.267	& - & - & .641 & - & - \\ 
        average & .192	&\textbf{.179}	&.187	& - & - & .482 & - & - \\ 
        \midrule
        
        & \multicolumn{8}{c}{LM: \texttt{LLaMA-13B}} \vspace{0.05cm} \\
        \hline
        MWP & .159	&\textbf{.117}	&.151	& - & - & .428 & - & - \\ 
        MHQA & .207	&\textbf{.142}	&.149	& - & - & .319 & - & - \\ 
        Plan. & .205	&\textbf{.147}	&.217	& - & - & .244 & - & - \\ 
        Relation.& \textbf{.239}	&.334	&.268	& - & - & .602 & - & - \\ 
        average & .189	&\textbf{.153}	&.170	& - & - & .391 & - & - \\ 
        \midrule
        
        & \multicolumn{8}{c}{LM: \texttt{LLaMA-70B}} \vspace{0.05cm} \\
        \hline
        MWP & .153	&\textbf{.083}	&.120	& - & - & .233 & - & - \\ 
        MHQA & .174	&\textbf{.082}	&.117	& - & - & .193 & - & - \\ 
        Plan. & .229	&\textbf{.156}	&.207	& - & - & .172 & - & - \\ 
        Relation.& \textbf{.223}	&.333	&.257		& - & - & .539 & - & - \\ 
        average & .176	&\textbf{.118}	&.144	& - & - & .247 & - & - \\ 
        \midrule
        
        & \multicolumn{8}{c}{LM: \texttt{Mistral-7B}} \vspace{0.05cm} \\
        \hline
        MWP & .165	&\textbf{.111}	&.138	& - & - & .331 & - & - \\ 
        MHQA & .184	&\textbf{.120}	&.135	& - & - & .271 & - & - \\ 
        Plan. & .260	&\textbf{.194}	&.231	& - & - & .198 & - & - \\ 
        Relation.& \textbf{.207}	&.295	&.245		& - & - & .561 & - & - \\ 
        average & .186	&\textbf{.144}	&.159	& - & - & .322 & - & - \\ 
        \midrule
        
        & \multicolumn{8}{c}{LM: \texttt{Mistral-7B-instruct}} \vspace{0.05cm} \\
        \hline
        MWP & .154	&\textbf{.108}	&.127	& - & - & .349 & - & - \\ 
        MHQA & .205	&.191	&\textbf{.186}	& - & - & .355 & - & - \\ 
        Plan. & .203	&\textbf{.155}	&.212	& - & - & .328 & - & - \\ 
        Relation.& \textbf{.238}	&.402	&.310		& - & - & .659 & - & - \\ 
        average & .186	&\textbf{.174}	&.177	& - & - & .383 & - & - \\ 
        
        \bottomrule
    \end{tabular}
    }
    \caption{ECE score ($\downarrow$) for all LMs on four domains -- Math Word Problems (MWP), Multi-hop QA (MHQA), Planning (Plan.), and Relational Inference (Relation.) -- averaged across five prompting strategies. The best score is \textbf{in bold}. }
    \label{table:results_ece}
\end{table*}

\end{document}